\documentclass[10pt,journal,compsoc]{IEEEtran}

\ifCLASSOPTIONcompsoc
  \usepackage[nocompress]{cite}
\else
  \usepackage{cite}
\fi

\usepackage[pdftex]{graphicx}

\usepackage[utf8]{inputenc} 
\usepackage[T1]{fontenc}    
\usepackage{hyperref}       
\usepackage{url}            
\usepackage{booktabs}       
\usepackage{amsfonts}       
\usepackage{nicefrac}       
\usepackage{microtype}      
\usepackage{xcolor}         
\usepackage{algorithm2e}
\usepackage{graphicx,wrapfig}
\usepackage{caption} 

\begin{document}

\title{TokenLearner: What Can 8 Learned Tokens Do for Images and Videos?}

%

\author{%
  Michael S. Ryoo, AJ Piergiovanni, Anurag Arnab, Mostafa Dehghani, Anelia Angelova\IEEEcompsocitemizethanks{\IEEEcompsocthanksitem M. Ryoo, A. Piergiovanni, A. Arnab, M. Dehghani, and A. Angelova are with Google Research.\protect\\
E-mail: \{mryoo, ajpiergi, aarnab, dehghani, anelia\}@google.com
\IEEEcompsocthanksitem M. Ryoo is also with Stony Brook University.}
\thanks{Manuscript received Feb 15, 2022.}
}


\IEEEtitleabstractindextext{%
\begin{abstract}
In this paper, we introduce a novel visual representation learning which relies on a handful of adaptively learned tokens, and which is applicable to both image and video understanding tasks. Instead of relying on hand-designed splitting strategies to obtain visual tokens and processing a large number of densely sampled patches for attention, our approach learns to mine important tokens in visual data. 
This results in efficiently and effectively finding a few important visual tokens and enables modeling of pairwise attention between such tokens, over a longer temporal horizon for videos, or the spatial content in images. 
Our experiments demonstrate strong performance on several challenging benchmarks for both image and video recognition tasks. Importantly, due to our tokens being adaptive, we accomplish competitive results at significantly reduced compute amount.
We obtain comparable results to the state-of-the-arts on ImageNet while being computationally more efficient. We also confirm the effectiveness of the approach on multiple video datasets, including Kinetics-400, Kinetics-600, Charades, and AViD.

The code is available at: \url{https://github.com/google-research/scenic/tree/main/scenic/projects/token_learner}
\end{abstract}

\begin{IEEEkeywords}
Computer vision, Activity recognition
\end{IEEEkeywords}}

\maketitle

\IEEEdisplaynontitleabstractindextext

\IEEEpeerreviewmaketitle

\IEEEraisesectionheading{\section{Introduction}}


Images and videos provide an abundance of visual information.
Image understanding is a long standing problem in computer vision, and despite incredible advances, obtaining the best visual representation for a variety of image understanding tasks is still an active area of research. 
Videos, in addition to addressing a similar image understanding task, require employing effective spatial-temporal processing of both RGB and time streams to capture long-range interactions~\cite{carreira2017quo,tran2014c3d,kay2017kinetics,hara2017learning,monfort2018moments,feichtenhofer2018slowfast,stroud2018d3d,3dconv,piergiovanni2020elo,alayrac2020selfsupervised}. An important aspect of this understanding is how to quickly learn which parts of the input video stream are important, both spatially and temporally, and to focus computational resources on them.
But what basic processing mechanism are able to do so successfully for both images and videos?



Transformer models using multi-head self-attention~\cite{vaswani2017attention} have been very successful in both image and video recognition. Extending its original usage for text, the Vision Transformers (ViT)~\cite{dosovitskiy2020} takes advantage of Transformers by treating an image as a sequence of patch tokens (e.g., 16x16 patches). At every layer, a ViT model recombines and processes patch tokens based on pairwise relations between the tokens, constructing a global representation of the entire image. The effectiveness of Transformers has also been shown in many computer vision tasks such as object detection~\cite{liu2021swin} and video classification~\cite{arnab2021vivit}.

The main challenge in many Vision Transformer architectures is that they often require too many tokens to obtain reasonable results. Even with 16x16 patch tokenization, for instance, a single 512x512 image corresponds to 1024 tokens. In the case of videos with multiple frames, this results in tens of thousands of tokens as each video `tubelets’ (e.g., 16x16x2 video segments) becomes a token. Further, such large number of tokens need to be processed at every layer, as the outputs from the previous layer become the tokens for the next layer. Considering that the Transformer computation (and memory) increases quadratically with the number of tokens, this can often make Transformers intractable for larger images and longer videos. This leads to the question: \emph{is it really necessary to process that many tokens at every layer?}

In this paper, we show that adaptively generating a smaller number of tokens, rather than always relying on tokens formed by uniform splitting, enables Vision Transformers to run much faster and perform better. TokenLearner is a learnable module that takes an image-like tensor (i.e., input) and generates a small set of tokens. This module could be placed at various different locations within the model of interest, significantly reducing the number of tokens to be handled in all subsequent layers. The experiments demonstrate that having TokenLearner saves memory and computation by half or more without damaging classification performance. Furthermore, because of its ability to adapt to inputs, it often is capable of increasing the recognition accuracy while relying on less amount of computation.



We formulate TokenLearner using a straightforward spatial attention mechanism. The idea is to learn to adaptively compute important regions in the input image/video, and generate tokens out of such regions. We compute spatial attention maps highlighting regions-of-importance (using convolutional layers or MLPs), and they are applied to the input itself to weight each region differently (and discard unnecessary regions). The results are spatially pooled to generate the final learned tokens. This is in contrast to previous approaches which densely sampled tokens e.g., 16x16 or 32x32 for either images or videos~\cite{dosovitskiy2020,bertasius2021timesformer}.


In our study, we find that very few tokens may be sufficient for a visual understanding task. More specifically, for images we show that one can significantly reduce the computational budget of the Vision Transformer, when learning 8-16 tokens as an intermediate representation (instead of keeping 200$\sim$500). We experimentally confirm that TokenLearner is able to reduce the number of total FLOPS, while maintaining or even increasing the classification accuracy. Similarly, for video recognition we show improved performance over the state-of-the art on three challenging datasets while only using 8-16 tokens per frame. 


The approach is simple, efficient, and, as shown by the results, outperforms methods including both convolutional methods and previous space-time Transformer baselines without TokenLearner. We demonstrate that our models with TokenLearner performs comparably to previous Transformer models on ImageNet (and ImageNet ReaL) while meaningfully reducing the computation. In video understanding tasks, TokenLearner established new state-of-the-art numbers on multiple challenging video datasets.

This paper extends an earlier version~\cite{ryoo2021tokenlearner_neurips} published at a conference, by generalizing the TokenLearner for both image and video representation learning. Unlike the conference version which only focused on videos, in this manuscript, we add an extensive amount of new experiments confirming the benefits of TokenLearner on both image and video classifications. It also includes various detailed ablation experiments with further analysis.

\section{TokenLearner Modules for Adaptive Tokenization}
\label{sec:main}

In vision transformer architectures such as ViT~\cite{dosovitskiy2020}, an input image is first tokenized by splitting it into small (e.g., 16x16) spatial patches, which are used as input to the model. Similarly, in recent video transformer architectures, such as ViViT~\cite{arnab2021vivit} and TimeSformer~\cite{bertasius2021timesformer}, the video is tokenized by cutting the video into 2d spatial or 3d spatio-temporal cubes on a regular grid.

Instead of processing fixed, tokenized inputs, our attention module learns the tokens that are to be used for the recognition task. 
We gain several important properties by doing so:  (1) We enable the adaptive tokenization so that the tokens can be dynamically selected conditioned on the input. (2) This also effectively reduces the total number of tokens for the transformer, which is particularly beneficial considering that there are many tokens in videos and the computation is quadratic to the number of tokens. 3) Finally, we provide an ability for each subsequent layer to learn to rely on different space-time tokenizations, potentially allowing different layers to capture different aspects of the video.
%
%
%
%
These dynamically and adaptively generated tokens can be used in standard transformer architectures such as ViT for images and ViViT for videos, or can be used within the specialized video architecture which we discuss further in Section~\ref{sec:for_videos}.






\subsection{TokenLearner}
\label{sec:token_learner}

Let $X$ be an input tensor with a space-time shape: $X \in \mathbb{R}^{T \times H \times W \times C}$ where $H \times W$ corresponds to the spatial dimension of the input, $T$ is the temporal dimension (i.e., number of frames), and $C$ is the number of channels. Let $X_t$ be a temporal slice of it, corresponding to the frame $t$: $X_t \in \mathbb{R}^{H \times W \times C}$. In the case of an image input, $T = 1$ and $X = X_t$. Note that $X$ could also be an intermediate representation within a network, and $X_t$ will be its slice in such case.

For every time frame $t$, we learn to generate a series of $S$ tokens, $Z_t = [z_i]_{i=1}^S$, from the input frame $X_t$. Specifically, we formulate a tokenizer function, $z_i = A_i(X_t)$, which maps the input frame $X^t$ to a token vector $z_i$: $\mathbb{R}^{H \times W \times C} \mapsto \mathbb{R}^C$. The idea is to learn our tokenizer function $A_i$ to adaptively select an informative combination of pixels (or spatial locations) in $X_t$, and we have $S$ number of such functions. This way, our tokens will not be fixed splits of the input tensor, but a set of adaptively changing spatial selections. Different tokens will be mined per frame, allowing us to model their space-time relations/interactions in case of videos. We also set $S$ to be smaller than $H \times W$ (e.g., $S=8$ and $H \times W = 32\times 32$), enabling the model to significantly reduce the computations needed for the layers following this module.



Here, our tokenizer $z_i = A_i(X_t)$ is implemented with a spatial attention mechanism: i.e., the model learns to compute a weight map (of size $H \times W$) conditioned on the input $X_t$, and is multiplied with $X_t$ itself.
More specifically, let $\alpha_i(X_t)$ be a function generating the spatial $H \times W \times 1$ weight map. Each token $z_i$ is generated by
\begin{equation}
    z_i = A_i(X_t) = \rho(X_t \odot A_{iw}) = \rho(X_t \odot \gamma(\alpha_i(X_t))),
\end{equation}
where $\odot$ is the Hadamard product (i.e., element-wise multiplication) and $A_{iw} \in \mathbb{R}^{H \times W \times C}$ is an intermediate weight tensor computed with the function $\alpha_i(X_t)$ and the broadcasting function $\gamma(\cdot)$. Finally, spatial global average pooling $\rho(\cdot)$ is applied on top of them to reduce the dimensionality to $\mathbb{R}^C$. The resulting tokens are gathered to form the output tensor: $Z_t = [z_i]_{i=1}^S \in \mathbb{R}^{S \times C}$.

The overall process has a form of an element-wise spatial self-attention. In our initial version, $\{\alpha_i(\cdot)\}_{i=1}^S$ are implemented together as a single or a series of convolutional layers (with the channel size $S$) followed by a sigmoid function. In our version 1.1, it is implemented with a single MLP layer (i.e., two dense layers with gelu in between). In case of an image, $Z = Z_t$.  In the case of a video, the tokens $Z_t$ from all the frames are collected to form the final output token tensor $Z \in \mathbb{R}^{ST \times C}$.

\begin{figure}
    \centering
    \includegraphics[width=0.99\linewidth]{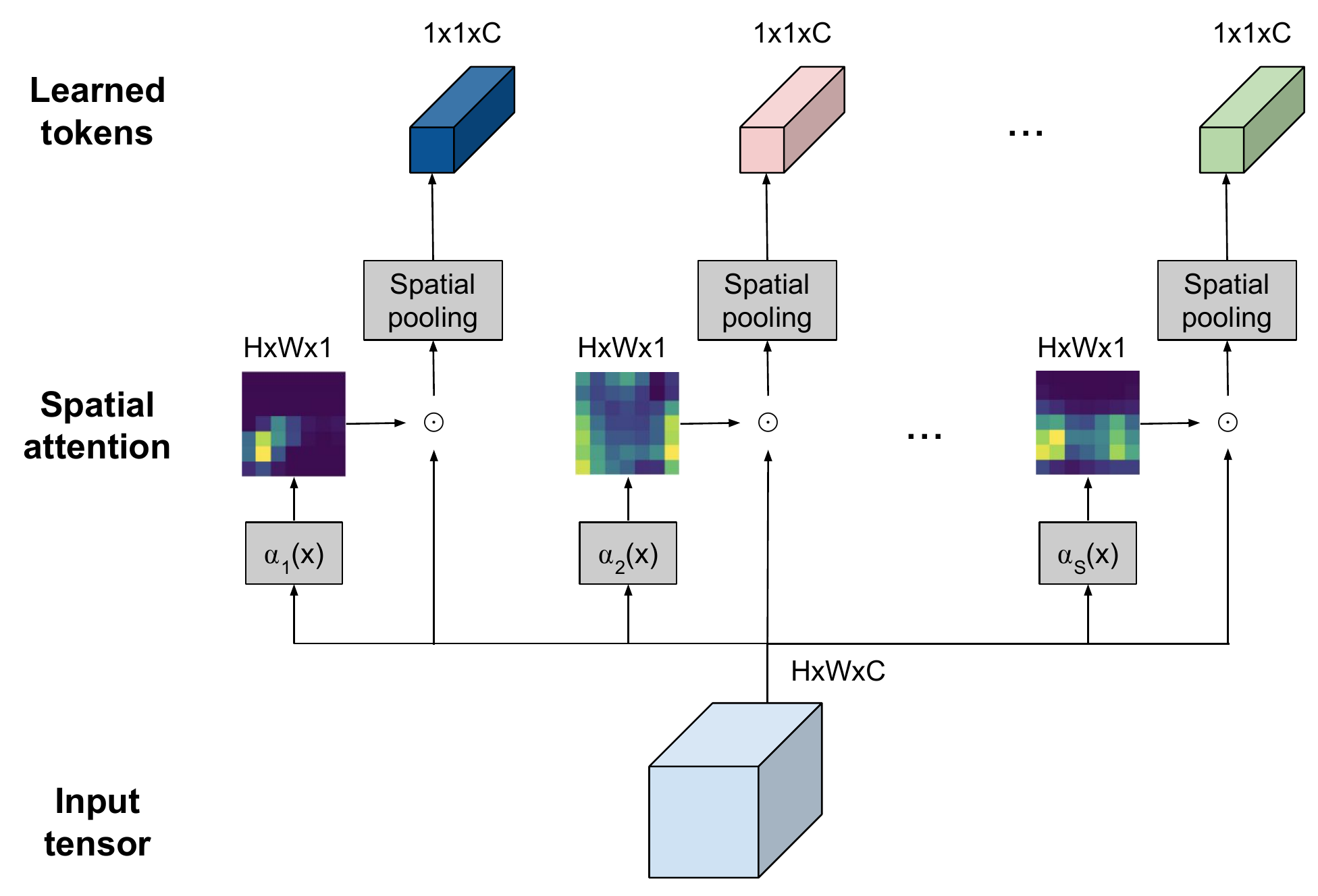}
    \caption{Visual illustration of the TokenLearner module, applied to a single image. TokenLearner learns to spatially attend over a subset of tensor pixels (i.e., from intermediate spatial representations), and generates a set of token vectors adaptive to the input.}
    \label{fig:module}
\end{figure}

We specifically name our token learning module as ``TokenLeaner''. Figure~\ref{fig:module} visually summarizes the TokenLearner module. 

\subsubsection{Compute reduction in Transformers:}
The learned tokens (i.e., the outputs of the TokenLearner $Z$) are provided to the subsequent layers for the visual representation learning, such as multi-head self-attention (MHSA) used in Vision Transformer and ViViT. With the TokenLearner, these subsequent layers only need to process a small number of tokens (e.g., 8 instead of 1024) and this significantly reduces the computations, as they are quadratic to the number of tokens. Figure \ref{fig:vit-tl} (a) shows a basic architecture inserting the TokenLearner module within ViT. It could be added at any location within the network, and the relative compute of the Transformer layers after the TokenLearner become almost negligible due to the huge difference in the number of tokens.


\subsection{TokenFuser}
\label{sec:fusion}

After the TokenLearner generates tokens and its subsequent Transformer layer (e.g., MHSA) processes them, the ``TokenFuser'' could be used to further (1) fuse information across the tokens and (2) remap the representation back to its original spatial resolution.
This enables the model to capture spatial (or spatio-temporal) `patterns' formulated by the tokens, and recover the original input tensor shape when necessary.


First, given the token tensor $Y \in \mathbb{R}^{ST \times C}$ from a Transformer layer, we apply a linear layer (i.e., a fully connected MLP layer) over the tokens, not channels.
That is, we learn a linear function of $\mathbb{R}^{ST} \mapsto \mathbb{R}^{ST}$
where $S$ is the number of our tokens mined per frame and $T$ is temporal size of the input tensor, and apply it to every channel independently. 
That is, we update $Y = (Y^T M)^T$ where $M$ is a learnable weight matrix with size $ST \times ST$. The result of such operation maintains the tensor size of $ST \times C$.
We believe this also has a connection to the observations from the concurrent work, MLPMixer~\cite{tolstikhin2021mlpmixer}, that it is beneficial to have token-wise linear layers capturing patterns formed by tokens.

Next, the TokenFuser processes each temporal slice $Y_t \in \mathbb{R}^{S \times C}$ individually, and remaps the token tensor of size $S \times C$ back to $H \times W \times C$, by learning to combine the tokens for each spatial location in $H \times W$ differently.
\begin{equation}
X_t^{j+1} = B(Y_t, X_t^j) = B_w Y_t + X_t^j = \beta_i(X_t^j) Y_t + X_t^j
\label{eq:fuser}
\end{equation}
where $X_t^j$ is the residual input to the previous TokenLearner module, $Y_t$ is the processed tokens in the TokenFuser module, and $X_t^{j+1}$ is the output. $B_w \in \mathbb{R}^{HW \times S}$ is an intermediate weight tensor computed with the function $\beta_i(X_t)$. The function $\beta_i(X_t)$ is implemented with a simple linear layer followed by a sigmoid function.

\begin{figure}
    \centering
    \includegraphics[width=0.99\linewidth]{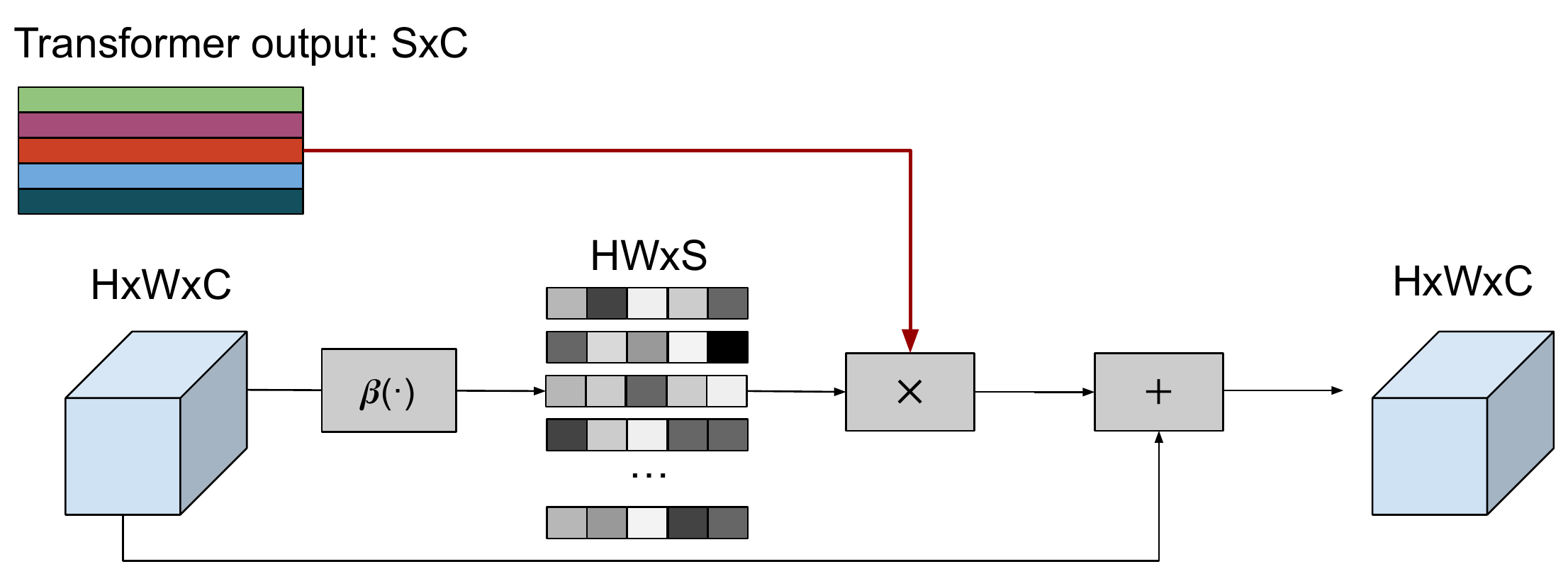}
    \caption{Visual illustration of the TokenFuser module, applied to each image frame individually.}
    \label{fig:fusermodule}
\end{figure}

Figure \ref{fig:fusermodule} illustrates the overall process of the TokenFuser (the token-wise linear layer is omitted).

\section{Experiments with Images}
\label{sec:exp-img}

In order to validate the power of the TokenLearner module, we first try TokenLearner on image representation learning.  We evaluate two different architectures: (a) simply inserting the TokenLearner within standard transformer models, and (b) using the TokenFuser in addition to the TokenLearner at multiple locations within the transformers.


\subsection{Network architecture implementation}
\label{subsec:image-network}

We use the Vision Transformer architecture \cite{dosovitskiy2020}, following its detailed settings and implementation \cite{dehghani2021scenic}. We use ViT-B/16 and ViT-L/16 as our backbone, while also applying the TokenLearner to ViT-B/32 (the same model but with an initial patch size of 32x32 in the beginning), ViT-S/32 (smaller version with 384 channels), and more.
The ViT-S and ViT-B backbone models have 12 transformer layers, while ViT-L has 24.
Following the exact setting of \cite{dosovitskiy2020}, we used the input resolution of 224x224, 384x384, or 512x512 depending on the dataset and the model (i.e., 196, 576, or 1024 tokens). Positional encodings identical to ViT are used.

\begin{figure}
    \centering
    \includegraphics[width=0.8\linewidth]{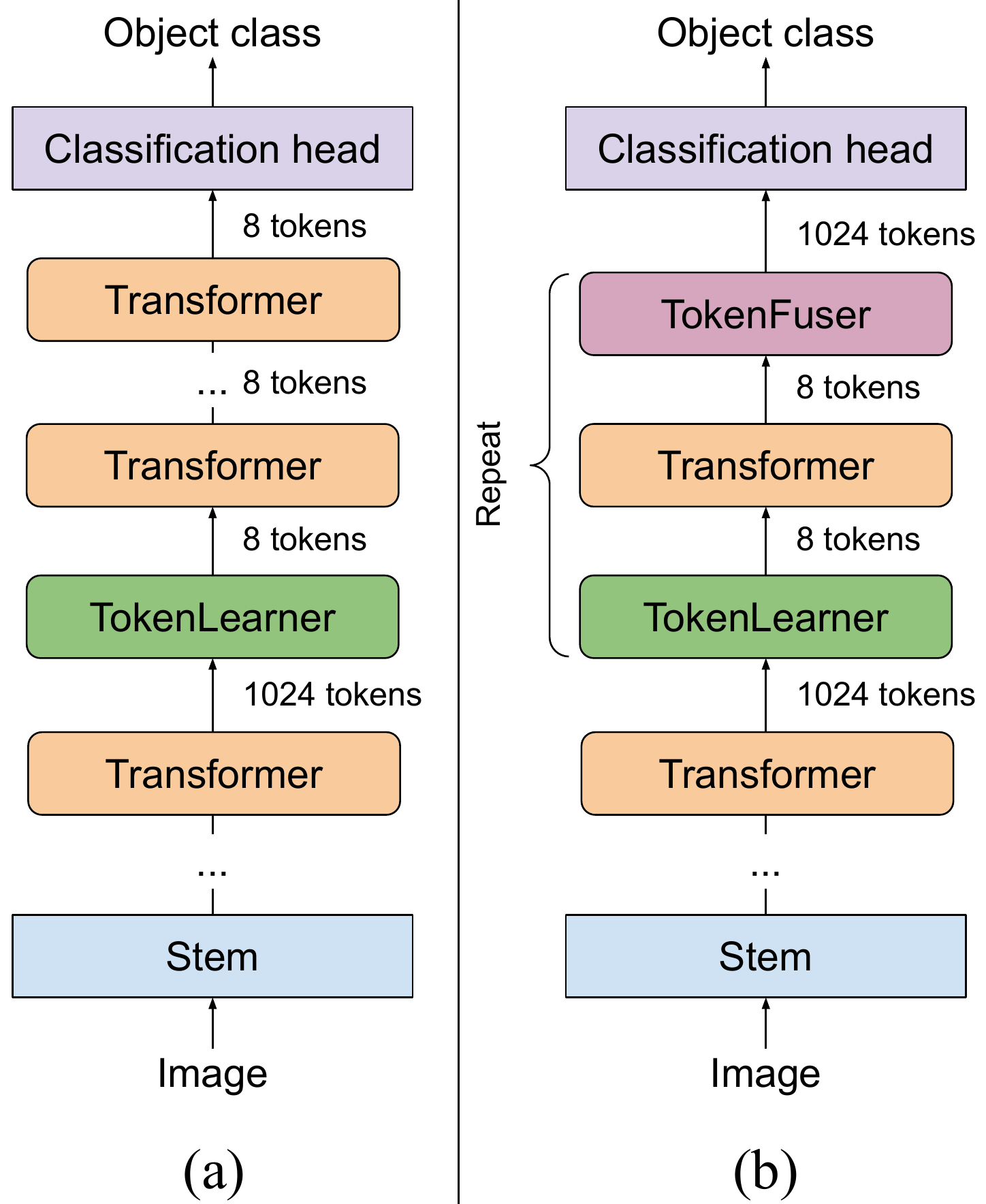}
    \caption{Our models following the ViT architecture. (a) with TokenLearner and (b) with both TokenLearner and TokenFuser.}
    \label{fig:vit-tl}
\end{figure}

Figure~\ref{fig:vit-tl} (a) and (b) show two different architectures incorporating TokenLearner. (a) is formed by inserting TokenLearner in the middle of the network such as after the 6th transformer among 12, while (b) uses both TokenLearner and TokenFuser. In particular, our model (b) is formed by replacing conventional Transformer layers with a series of TokenLearner-Transformer-TokenFuser. Similar to (a), such replacement is done only for the layers after a certain layer. For instance, we keep six of the standard Transformer MHSA layers in the beginning, and replaces the remaining six layers with our TokenLearner-Transformer-TokenFuser modules repeated six times. 
We also modified some of our models to have more transformer layers (e.g., 21 instead of 12), and we specify this when we do so. Note that the computation increase caused by the transformer layers added after TokenLearner module is very small, as the number of tokens in these layers are few: 8 or 16.

We tried various number of tokens including $S=8, 16, 32$, and use $S=8$ and $16$ as our default settings. That is, the TokenLearner is learning to abstract an image into 8 (or 16) tokens. The spatial attention function ($\alpha$) in TokenLearner is implemented with four 3x3 conv. layers (with gelu in between), whose channel size is identical to the number of tokens (e.g., $S=8$). 

We adopt the training settings (e.g., learning rate, training epochs, etc.) of \cite{dosovitskiy2020}.


\subsection{Image classification datasets}
\label{sec:image_datasets}

\begin{figure*}
    \centering
    \includegraphics[width=0.49\linewidth]{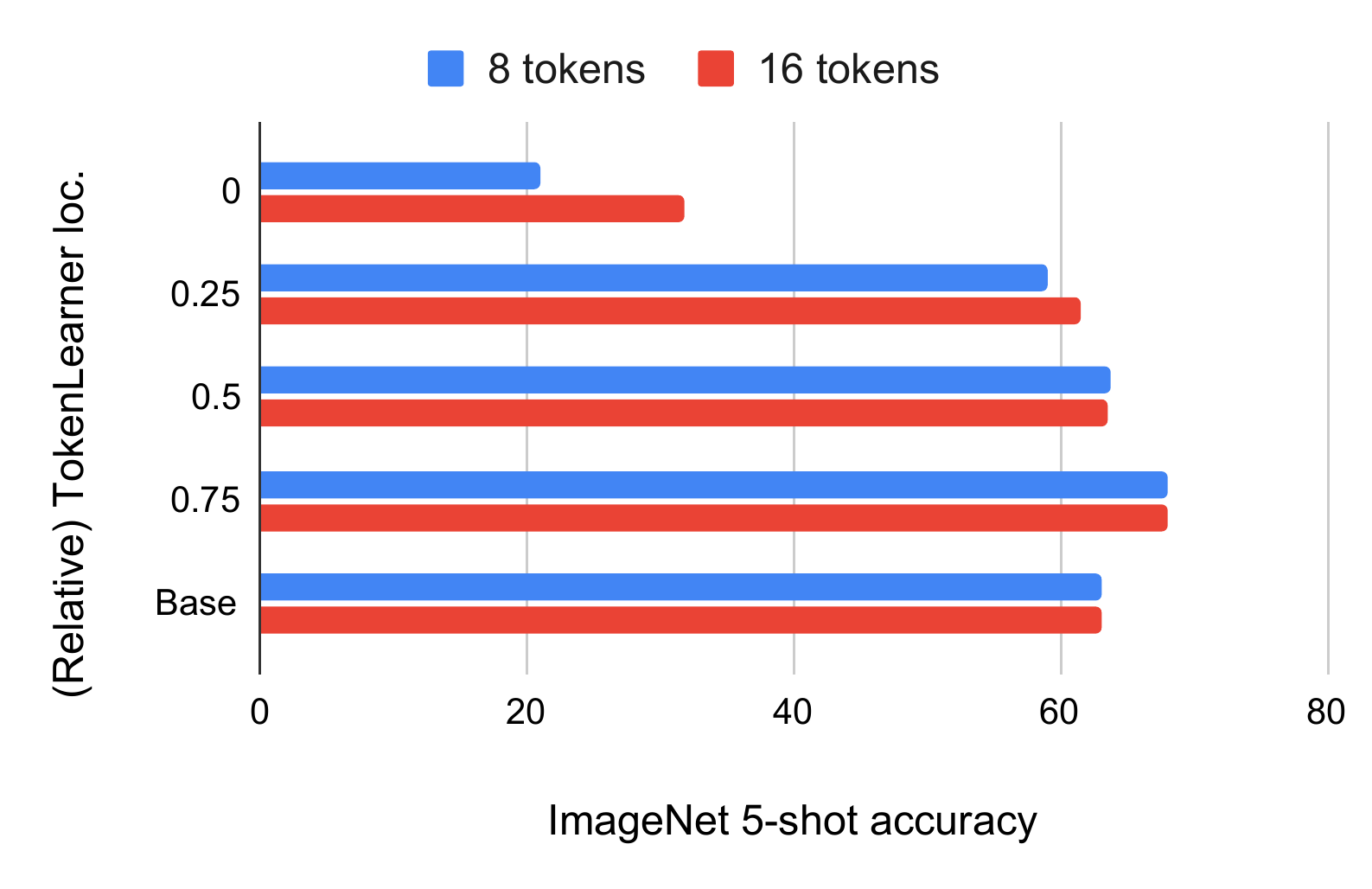}
    \includegraphics[width=0.49\linewidth]{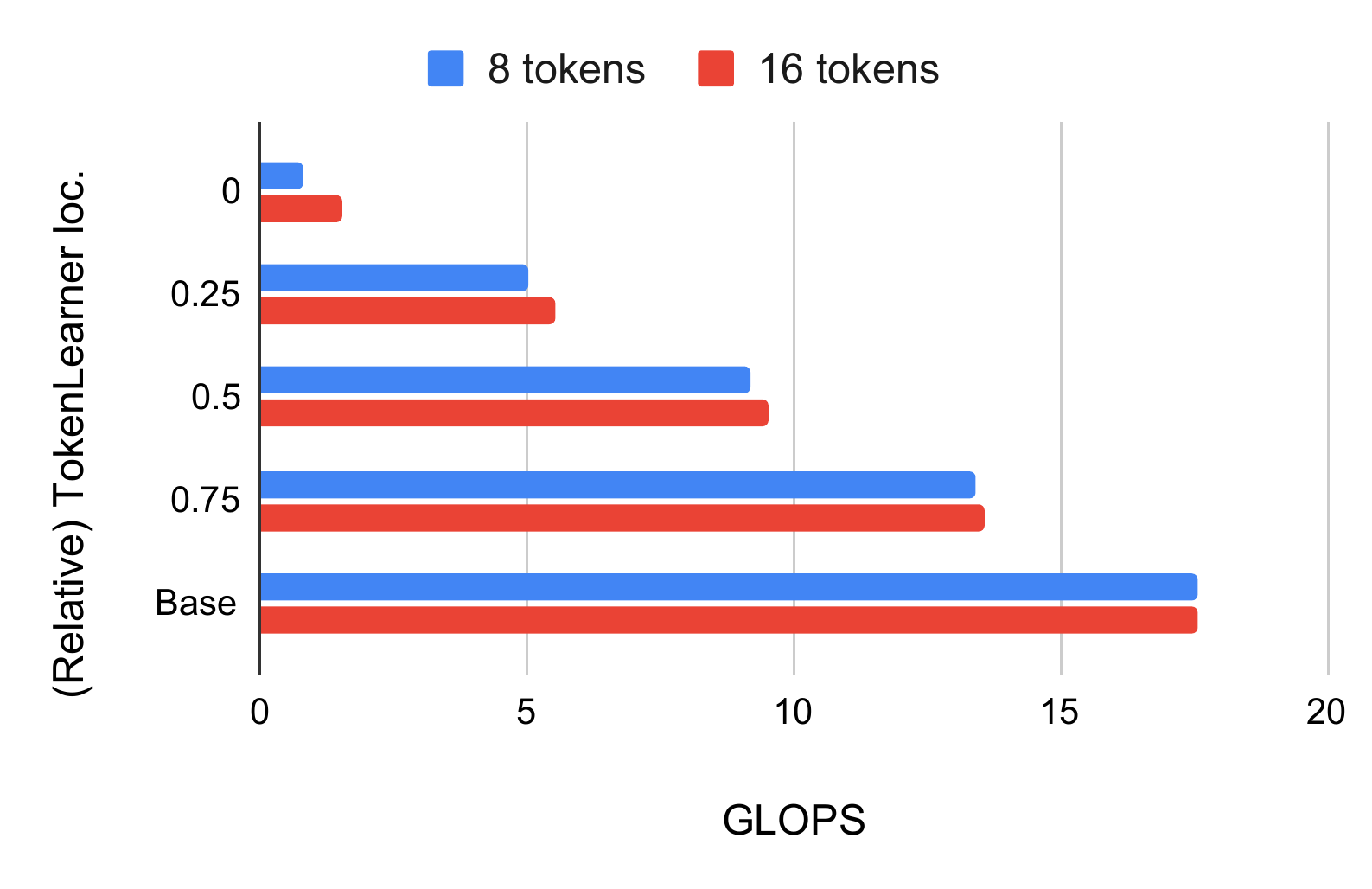}
    \caption{ImageNet 5-shot accuracy (left) and FLOPS (right) per different TokenLearner location within the model. `0' means that the TokenLearner is at the very beginning of the model (before any transformer), `0.5' means the middle of the model, and `Base' means that there is no token learning.}
    \label{fig:location}
\end{figure*}

\textbf{ImageNet}: We use the popular image benchmark, ImageNet~\cite{deng2009imagenet}. For our experiments, we use  the standard ImageNet version which has 1000 categories and 1.1M images. We use the image resolution of 384x384 for S/16 and B/16 models, and use 512x512 for L/16 models.
ImageNet ReaL \cite{beyer2020imagenet}, which is the dataset with Re-Assessed (ReaL) labels for ImageNet, was also used for the evaluation.

\textbf{JFT-300M}. The JFT-300M dataset is an internal dataset collected for training image classification models, which was first introduced by~\cite{chen2017jft}. Images are harvested from the web and are filtered to maximize label precision. It contains 300M images and has been shown to be suitable for learning high-capacity models, such as transformers.

In this work, we use the JFT-300M dataset only for pre-training purposes, following the evaluation protocol, previously established for ViT~\cite{dosovitskiy2020}. We use the image resolution of 224x224 for this.

\subsection{Ablation: where should we have TokenLearner?}

We first conducted an ablation to decide the best location to place the TokenLearner within the model. Figure \ref{fig:location} shows the results. The experiment was done with our model (a), without TokenFuser. It is showing the few-shot classification accuracies on ImageNet with JFT pre-training, following the protocol of ViT~\cite{dosovitskiy2020}. 
In addition, we show how the computation amount (FLOPS) changes per TokenLearner location. Basically, due to the large difference between the number of tokens with and without the TokenLearner (e.g., 8 with TokenLearner vs. 196 without), the computation of the transformers after the TokenLearner module becomes almost negligible compared to the transformers before the TokenLearner location.

We found that inserting TokenLearner in the middle of the network (at 1/2) achieves almost identical accuracies, while cutting the computation by (almost) half. In addition, having the TokenLearner at the later layer (after 3/4 of the network) achieves even superior performance compared to not using the TokenLearner while performing faster, thanks to its adaptiveness.

\subsection{Results}
\label{sec:exp_image_results}

Following the protocol established in ViT~\cite{dosovitskiy2020}, we evaluated the models with and without TokenLearner in terms of (i) fine-tuning accuracies and (ii) few-shot accuracies. For the fine-tuning accuracies, we pre-train the model with JFT and fine-tune it with the original ImageNet dataset using an image resolution of 384x384 (for ViT-S and ViT-B models) or 512x512 (for ViT-L models) as done in previous works. For the few-shot accuracies, we also follow the protocol of ViT~\cite{dosovitskiy2020} where we do a regularized least-squares regression that maps the (frozen) representation of a subset of training images to $\{-1, 1\}^K$ target vectors. 
We report 5-shot ImageNet Top-1 accuracy, as well as the 5-shot accuracies averaged over multiple datasets: Caltech101, Caltech-UCSD Birds 2011, Cars196, CIFAR100, colorectal\_histology, DTD, ImageNet, Oxford-IIIT Pet, and UC Merced Land Use Dataset. Few-shot accuracies are particularly interesting as it shows the generalization capability of the representation itself being learned. We note that 5-shot accuracy was also used to evaluate the representations learned by ViT~\cite{dosovitskiy2020}. In the experiments in this subsection, we use the model (a) from Figure \ref{fig:location} which is without TokenFuser. We inserted the TokenLearner module exactly at the mid-point of the network unless specified otherwise. The default number of tokens was 8 and we also use 16, as they showed the best accuracy-speed balance.

\begin{figure}
    \centering
    \includegraphics[width=0.99\linewidth]{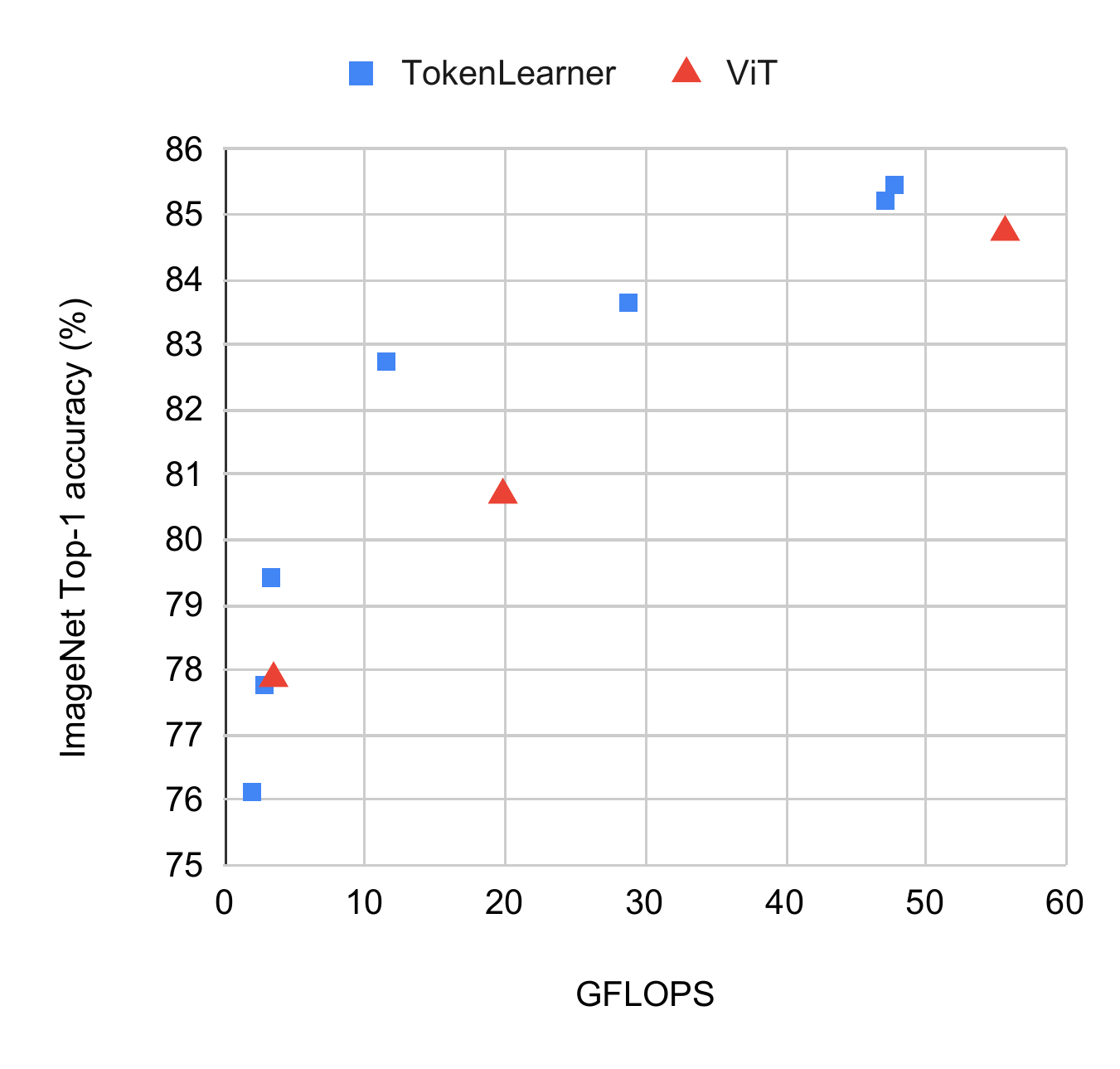}
    \caption{Visualization of ImageNet fine-tuning accuracies of the baseline ViT models vs. TokenLearner. X-axis is GFLOPs, which measures the amount of computation required.}
    \label{fig:imagenet-results}
\end{figure}

Figure \ref{fig:imagenet-results} and Table \ref{table-imagenet} shows the ImageNet fine-tuning evaluation results, using smaller ViT-S and ViT-B models. We show accuracies of various versions of ViT and their TokenLearner versions, as specified in Section~\ref{subsec:image-network}. We are able to observe that there is a substantial improvement in efficiency-accuracy trade-offs. 

\begin{table}
    \centering
    \caption{ImageNet fine-tuning Top1 accuracies and FLOPS. The numbers in the parenthesis are the number of transformer layers. 16-TokenLearner is with 16 tokens instead of 8.}
    \label{table-imagenet}
    \begin{tabular}{lll}
        \toprule
        Method & GFLOPS & Accuracy \\
        \midrule
        ViT S/32 & 3.4 & 77.87\\
        ViT B/32 & 19.8 & 80.69\\
        ViT B/16 & 55.6 & 84.73\\
        \midrule
        TokenLearner S/32 & 1.9 & 76.13\\
        TokenLearner B/16 & 28.7 & 83.65\\
        \midrule
        TokenLearner S/32 (22) & 3.3 & 79.42 \\
        TokenLearner B/32 (20) & 11.5 & 82.74\\
        TokenLearner B/16 (21) & 47.1 & 85.21\\
        16-TokenLearner B/16 (21) & 47.7 & 85.45\\
        \bottomrule
    \end{tabular}
\end{table}

When directly applied to ViT (e.g., B/16), TokenLearner maintains the accuracy of ViT, while reducing the computation by almost a half. When more layers are used together with the TokenLearner, we are able to utilize the computation saved by the TokenLearner in the form of additional layers, and it performs superior while still having less FLOPS. The number of tokens the baseline ViT B/16 processes is 576, while the TokenLearner learns $S=8$ tokens. As a result, as mentioned in the above subsection, the computation of the transformers after the TokenLearner module becomes almost negligible compared to the transformers before the TokenLearner location.

\begin{figure}
    \centering
    \includegraphics[width=0.49\linewidth]{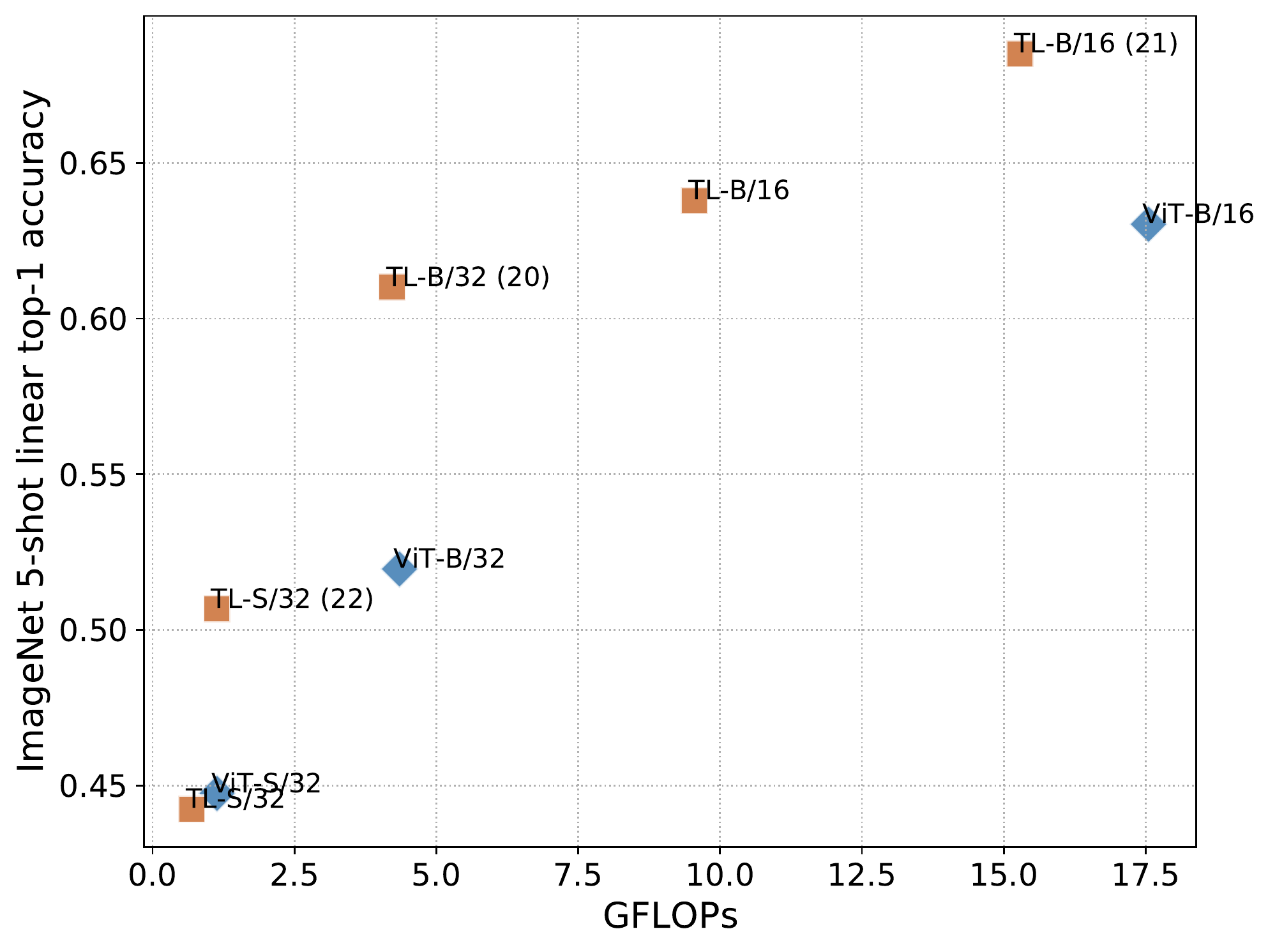}
    \includegraphics[width=0.49\linewidth]{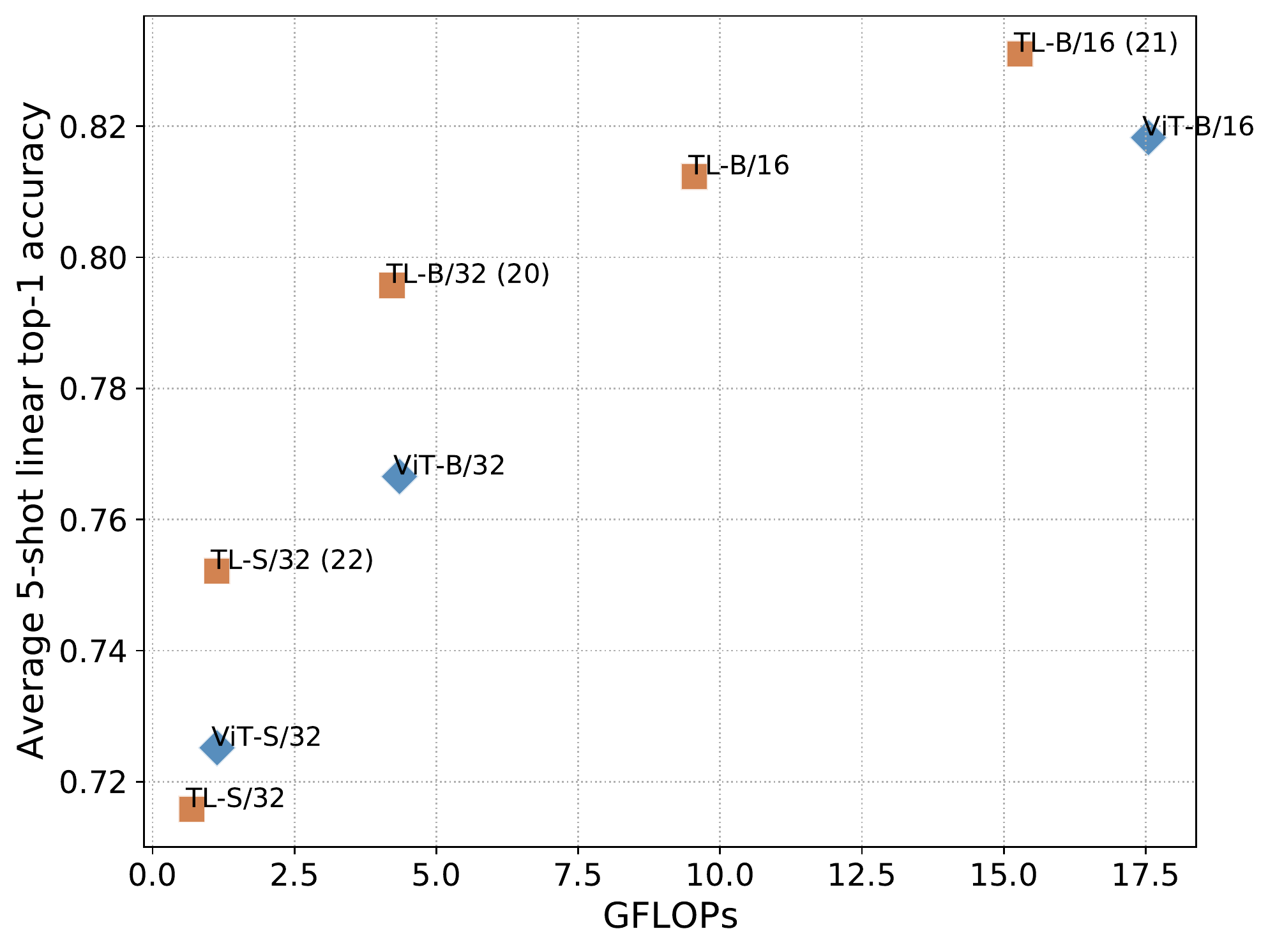}
    \caption{Few-shot classification experiments. It shows 5-shot classification accuracies on ImageNet (left) and average of multiple datasets listed in Sec.~\ref{sec:exp_image_results} (right).
    `TL' stands for TokenLearner. Check Appendix for results with more models.}
    \label{fig:fewshot}
\end{figure}

Figure~\ref{fig:fewshot} shows few-shot experiment results. For these experiments, an image resolution of 224x224 is used following~\cite{dosovitskiy2020}. The baseline ViT therefore uses 196 tokens (as opposed to 576 used in ImageNet fine-tuning experiments). This makes the gap between the number of tokens used in TokenLearner and ViT smaller (compared to fine-tuning setting), increasing TokenLearner's relative accuracy. 
It is interesting to observe that the accuracies of TokenLearner do not drop (e.g., TokenLearner-B/16 vs. ViT-B/16), despite the difference in the number of tokens.

\subsubsection{TokenLearner on larger models.}

We also evaluated our TokenLearner inserted into a `large' model: ViT-L. In addition to the standard L/16 model that splits the scene into 16x16 patches, the same model with finer tokens were used including L/14, L/10, and L/8. Note that the model size stays the same, with 464M parameters, and only the inputs change for these models. As discussed above, the number of tokens become 8 or 16 after the TokenLearner layer, regardless the model. For these large models, the image resolution of 512x12 was used for ImageNet.


Table \ref{table-large} shows the results. It specifies how many tokens were used for each model as well as the location of the TokenLearner layer. ``16-TL at 12'' means the number of tokens were 16, and TokenLearner was inserted after the 12th Transformer layer (i.e., in the middle). Some of the models were set to use additional layers (e.g., `+11'), but the increase in FLOPS were negligible due to the number of tokens being small after the TokenLearner layer.

We are able to observe that, similar to our experiments with ViT-S and ViT-B, TokenLearner is able to save the amount of computations by half when inserted in the middle. Further, it is able to do so without sacrificing the accuracy, due to the adaptiveness in the token generation. When the saved compute was used in other forms (e.g., use of L/14 instead of L/16), it showed that it is able to meaningfully outperform the base model. The actual runtime of the base ViT L/16 vs. L/16 + TokenLearner was 1400 vs. 2000 images per second. It is not exactly linear to FLOPS due to the existence of other bottlenecks such as data loading.

Table \ref{table-sota} compares our models with TokenLearner against the larger ViT models from \cite{dosovitskiy2020} and \cite{zhai2021scaling}. Despite using much smaller number of parameters, our TokenLearner models perform comparably to the huge and giant ViT models.

\begin{table}
  \caption{TokenLearner with ViT L/16 and L/14. 512x512 input images used.}
  \label{table-large}
  \centering
  \begin{tabular}{lll|cc}
    \toprule
    Base  & \# layers & TokenLearner\ & GFLOPS & ImageNet Top1\\
    \midrule
    ViT L/16 & 24 & - & 363.1 & 87.35\\
    \midrule
    ViT L/16 & 24 & 16-TL at 12 & 178.1 & 87.68\\
    ViT L/16 & 24+11 & 16-TL at 12 & 186.8 & 87.47\\
    ViT L/16 & 24+6 & 8-TL at 18 & 274.2 & 88.11\\
    ViT L/14 & 24+11 & 16-TL at 18 & 361.6 & 88.37\\
    \bottomrule
  \end{tabular}
\end{table}


\begin{table}
  \caption{Comparison to state-of-the-art ViT models.}
  \label{table-sota}
  \centering
  \begin{tabular}{llcc}
    \toprule
    Method  & \# params. & ImageNet & ImageNet ReaL\\
    \midrule
    BiT-L & 928M & 87.54 & 90.54\\
    ViT-H/14 & 654M & 88.55 & 90.72\\
    ViT-G/14 & 1843M & \textbf{90.45} & 90.81\\
    \midrule
    TL L/10 (24+11) & \textbf{460M} & 88.5 & 90.75\\
    TL L/8 (24+11) & \textbf{460M} & 88.87 & \textbf{91.05}\\
    \bottomrule
  \end{tabular}
\end{table}

\subsubsection{Making larger models much more efficient}

We also confirmed the strategy of using TokenLearner at much earlier in the network, such as after the 2nd or 3rd attention layer. This makes the overall model even more computationally efficient than the smaller base ViT model (B/16), while performing better. Table \ref{table-large-early} shows the results. We are able to observe that, for instance, the L/16 model with TokenLearner at the 3th attention layer gets a superior accuracy than ViT B/16, while its run time is around half of B/16.

\begin{table}
  \caption{TokenLearner inserted earlier within ViT L/16. 384x384 input images used.}
  \label{table-large-early}
  \centering
  \begin{tabular}{lll|cc}
    \toprule
    Base  & \# layers & TokenLearner\ & GFLOPS & ImageNet Top1\\
    \midrule
    ViT B/16 & 12 & - & 55.63 & 84.73\\
    \midrule
    ViT L/16 & 24 & 16-TL at 2 & 20.91 & 83.89\\
    ViT L/16 & 24 & 16-TL at 3 & 28.66 & 85.40\\
    ViT L/16 & 24 & 16-TL at 6 & 51.92 & 86.44\\
    \bottomrule
  \end{tabular}
\end{table}

\subsection{Ablations and Comparisons}
\label{sec:image_ablationss}

\subsubsection{TokenFuser}

First, we compare the TokenLearner models with and without the TokenFuser module. More specifically, we compared the model (a) and the model (b) from Figure \ref{fig:location}, to confirm the effectiveness of the TokenFuser. Table \ref{table-fuser} shows the results.

\begin{table*}
  \caption{Models with TokenLearner, with and without TokenFuser. The model without TokenFuser is described in Figure \ref{fig:location} (a). The model with TokenFuser uses the architecture of Figure \ref{fig:location} (b).}
  \label{table-fuser}
  \centering
  \begin{tabular}{llll|ccc}
    \toprule
        Base  & \# layers & TokenLearner & TokenFuser & ImageNet Top1 & ImageNet ReaL & GFLOPS\\
        \midrule
        B/16 & 12 & 8-TL at 6 & N & 83.2 & 88.1 & 28.3\\
        B/16 & 12 & 8-TL at 6 & Y & 83.7 & 88.4 & 28.5 \\
        \midrule
        B/16 & 12 & 16-TL at 6 & N & 83.2 & 88.0 & 28.7\\
        B/16 & 12 & 16-TL at 6 & Y & 83.9 & 88.7 & 29.1 \\
        \midrule
        L/16 & 24 & 16-TL at 12 & N & 87.6 & 90.4 & 184.6\\
        L/16 & 24 & 16-TL at 12 & Y & 87.6 & 90.5 & 187.1 \\
        \midrule
        L/16 & 24 & 8-TL at 18 & N & 87.9 & 90.8 & 273.2\\
        L/16 & 24 & 8-TL at 18 & Y & 88.2 & 90.9 & 273.8\\
        \midrule
        L/10 & 24+11 & 16-TL at 18 & N & 88.5 & 90.7 & 849.0\\
        L/10 & 24+11 & 16-TL at 18 & Y & 88.5 & 90.9 & 856.9\\
        \bottomrule
  \end{tabular}
\end{table*}

\subsubsection{TokenLearner vs. pooling}

A straightforward alternative to the TokenLearner module is the use of spatial pooling to reduce the number of tokens. It can be done by spatially rearranging the tokens to have the height and width, and then applying conventional spatial pooling. This is similar to the pooling-based MHSA module used in \cite{fan2021multiscale}.

Table \ref{table-pooling} compares the TokenLearner against the spatial pooling. In all these experiments, ViT L/16 model was used. We are able to observe that there is a benefit in token `learning'. The pooling-based token reduction does have computation similar to the TokenLearner, but it loses its accuracy compared to the base model. On the other hand, TokenLearner performs a bit better than the base model despite the low computation.

\begin{table}
    \caption{TokenLearner compared against pooling-based token reduction.}
    \label{table-pooling}
    \centering
        \begin{tabular}{ll|cc}
            \toprule
            Details & ImageNet & GFLOPS\\
            \midrule
            Base ViT L/16 &  87.35 & 363.1\\
            \midrule
            2x2 pool at 9 and 18 &  85.63 & 144.3\\
            2x2 pool at 12 and 18 &  86.41 & 187.2\\
            4x4 pool at 12 & 83.93 & 184.4\\
            \midrule
            16-TL at 12 & 87.68 & 184.6\\
            \bottomrule
        \end{tabular}
\end{table}

\begin{figure}
    \centering
    \includegraphics[width=0.9\linewidth]{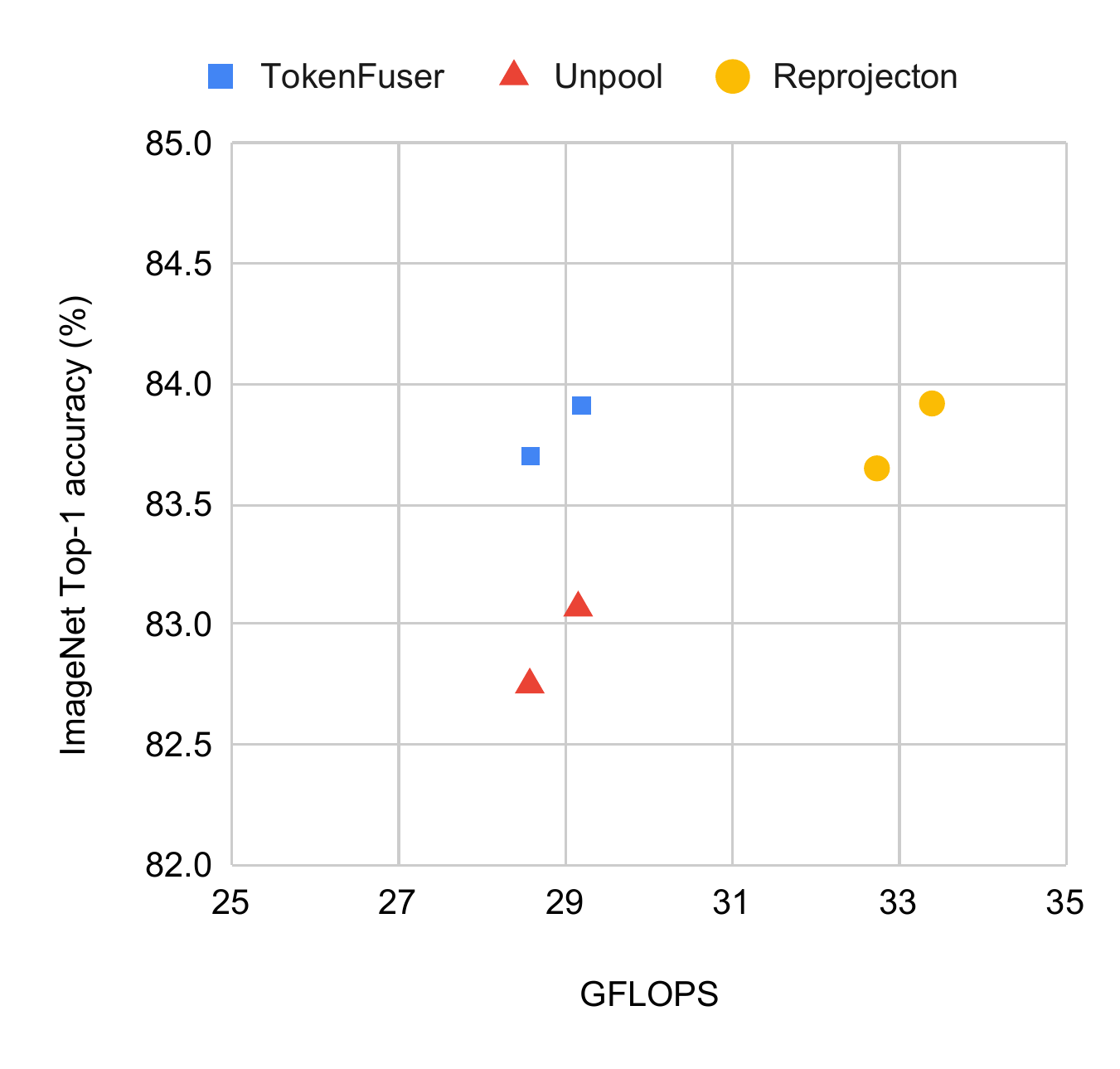}
    \caption{Ablations with TokenFuser alternatives.}
    \label{fig:fuser-ablation}
\end{figure}

\subsubsection{TokenFuser alternatives}

Here, we experimentally compare the proposed TokenFuser module with its alternatives. The role of the TokenFuser is to mix the output tokens from the Transformer layer and map it back to the original shape before the token reduction.

The most straightforward alternative would be to (1) use the masks from the TokenLearner module to `unpool' the output tokens. The idea is to multiply each output token with the corresponding spatial map computed during the previous TokenLearner module, and sum all of them to recover the original input tensor shape. Alternatively, (2) we can use one more transformer layer to increase the number of tokens back to the original number of tokens, similar to the `re-projection' used in \cite{wu2020visual}.

Figure \ref{fig:fuser-ablation} shows the results with B/16. The unpooling strategy performed worse. The reprojection strategy performed comparably to the TokenFuser, but required more FLOPS.

\section{TokenLearner for Videos}
\label{sec:for_videos}

\begin{figure*}
    \centering
    \includegraphics[width=0.9\linewidth]{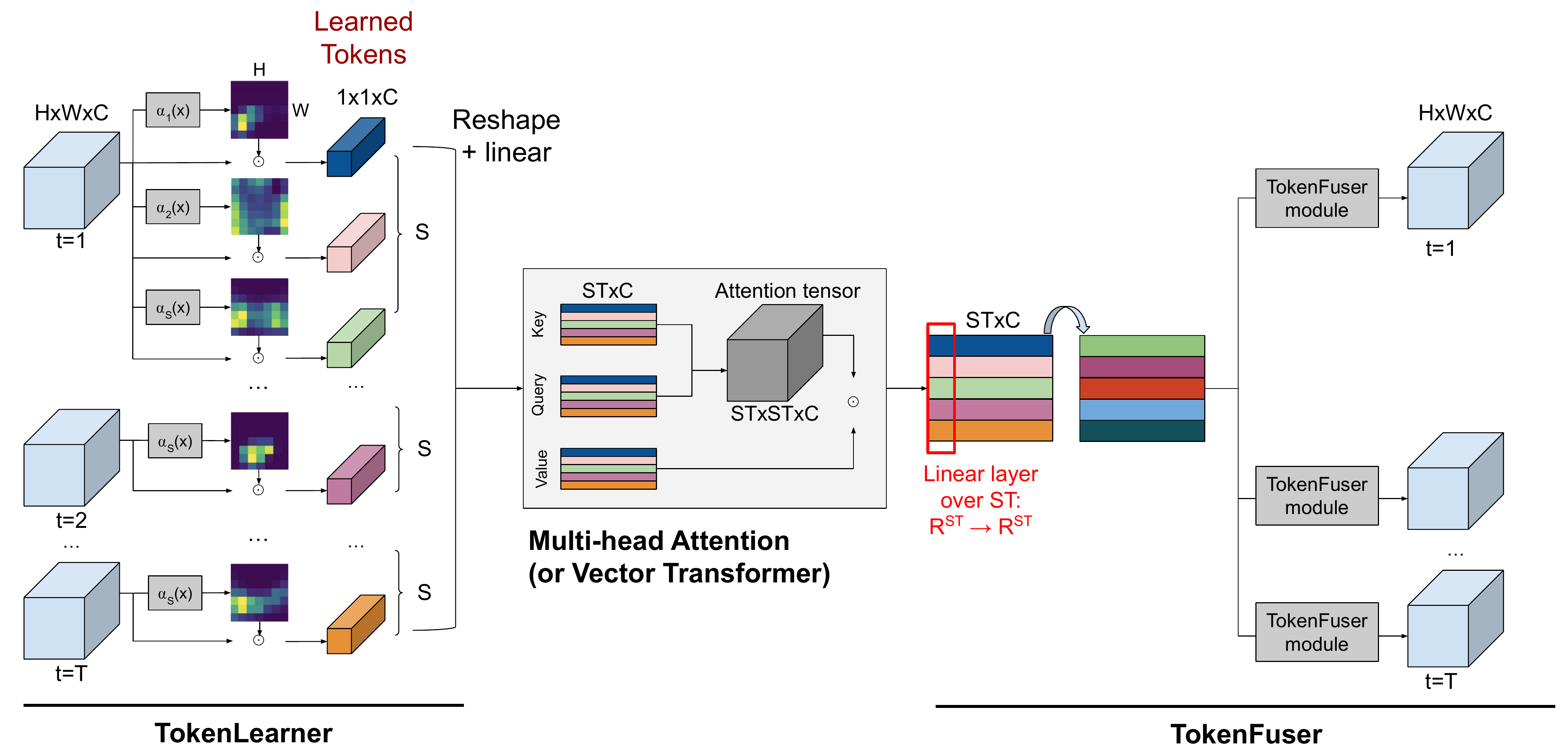}
    \caption{An illustration of TokenLearner, Transformer, and TokenFuser combined for video representation learning. TokenLearner first learns to generate a set of token vectors, the Transformer (e.g., MHSA) models their space-time relations, and TokenFuser combines them. $S$ is the number of tokens we learn per frame, and $T$ is the number of frames. $C$ is the number of channels, which we made to be identical across the modules for efficiency. Note that this combination can serve as a `module' itself, and one may stack such module multiple times within the network. TokenFuser could be dropped depending on the architecture.
    }
    \label{fig:main}
\end{figure*}

In this section, we illustrate how TokenLearner works for video representations.
The TokenLearner and TokenFuser modules introduced in Section \ref{sec:main} are directly applicable for video representation learning. The only difference between the TokenLearner used for images and it used for videos is that TokenLearner generates multiple $Z_t$ for videos and they need to be stacked to form $Z$. Once $Z$ is generated, any standard Transformer layers could be used to parse them jointly.

Figure~\ref{fig:main} provides an overview of the combined framework for videos. 
TokenLearner first extracts $S$ number of tokens per frame, resulting in a total of $ST$ tokens where $T$ is the number of frames. Once TokenLearner generates these adaptively learned tokens, they are provided to the subsequent Transformer layer to capture the global space-time patterns.
Finally (and optionally depending on the architecture), TokenFuser applies a linear layer over the token axis and then remaps the tensor shape back, as discussed in Section~\ref{sec:fusion}. Following Eq.~\ref{eq:fuser}, TokenFuser is applied for per-frame representation $Y_t$.
This results in a lightweight approach, which brings forth an efficient video representation by capturing long-range visual patterns.

What we show for the video representation learning is a combination of TokenLearner, Transformer, and TokenFuser modules repeated multiple times, as described in Figure~\ref{fig:vit-tl} (b). The TokenFuser part is dropped if we are using the model architecture (a), and only the Transformer layers are repeated multiple times after the TokenLearner module.

\begin{table*}
  \caption{Comparison of ViViT models with and without TokenLearner on Kinetics-400. GLOPS are per view. The difference in the number of parameters between the TokenLearner models (which are from Tables \ref{table-large} and \ref{table-fuser}) comes from the different number of layers used after the TokenLearner module.}
  \label{table-vivit-kin400}
  \centering
  \begin{tabular}{l|cccc}
        \toprule
        Method & Top-1 accuracy & Top-5 accuracy & \# params. & GFLOPS \\
        \midrule
        ViViT-L/16 \cite{arnab2021vivit} & 82.8 & 95.5 & 308M & 1446 \\
        ViViT-L/16 320 \cite{arnab2021vivit} & 83.5 & 95.5 & 308M & 3992 \\
        ViViT-H/14 \cite{arnab2021vivit} & 84.8 & 95.8 & 654M & 3981 \\
        \midrule
        ViViT-L/16 (our run) & 83.4 & 95.6 & 308M & 1446 \\
        \midrule
        TokenLearner 16at12 + L/16 & 83.5 & 95.6 & 308M & 766 \\
        TokenLearner 8at18 + L/16 & 84.5 & 96.1 & 383M & 1105 \\
        TokenLearner 16at18+ L/14 & 84.7 & 96.1 & 447M & 1621 \\
        TokenLearner 16at18+ L/10 & 85.4 & 96.3 & 450M & 4076 \\
        \bottomrule
  \end{tabular}
\end{table*}

\section{Experiments with Videos: TokenLearner with Video Vision Transformer}

\subsection{Network architecture implementation}

ViViT \cite{arnab2021vivit} is a direct extension of ViT for videos, which uses spatio-temporal patches from videos as its tokens. The size of the space-time patches are typically 16x16x2, which are given to the Transformer layers similar to ViT. ViViT and ViT share the architecture. For our experiments, we insert the TokenLearner module within the ViViT architecture, identically to how we inserted it within ViT in Figure~\ref{fig:vit-tl}. ViViT has been one of the state-of-the-art approaches for the Kinetics datasets \cite{carreira2017quo}, and the idea is to confirm whether TokenLearner could directly be added to such general video representation models and outperform them.

\subsection{Datasets and training}

For the experiments in this section, we use the Kinetics datasets, which are video classification datasets with relatively short video clips ($\sim$10 seconds). We train and evaluate on both Kinetics-400 and Kinetics-600 datasets, which have about 240k and 390k training samples. We follow the standard settings used in previous papers and report accuracy on the validation set \cite{carreira2017quo,feichtenhofer2018slowfast}.

Following ViViT \cite{arnab2021vivit}, we first pretrain models on JFT to obtain initial weights. We directly use the models from Section \ref{sec:exp-img}. The weights of the initial convolutional layers to handle image patches (e.g., 16x16) are processed to handle 16x16x2 video patches by following the 3D initialization strategies of ViViT, and the weights of the Transformer and the TokenLearner layers were directly adopted for the initialization. Next, the model was finetuned on the Kinetics data.


Similar to ViViT, we train the model for 30 epochs with the base learning rate of 0.05 with the Momentum optimizer.
Basically, all the settings in our Kinetics experiments follow the setting of ViViT.


\begin{table}
  \caption{ViViT + TokenLearner on Kinetics-400, compared to the previous models. Different approaches rely on different pre-training datasets, such as ImageNet-21K (for TimeSformer and Swin) and JFT (for ViViT and TokenLearner). The multiplication in GFLOPS correponds to the number of views used for the inference, such as 4x3 = 12.}
  \label{table-vivit-kin-sota}
  \centering
  \begin{tabular}{l|cc}
        \toprule
        Method & Top-1 accuracy & total GFLOPS \\
        \midrule
        R(2+1)D \cite{tran2018closer} & 73.9 & 304 $\times$ 115\\
        SlowFast 16x8, R101+NL \cite{feichtenhofer2018slowfast} & 79.8 & 234 $\times$ 30 \\
        TimeSformer-L \cite{bertasius2021timesformer} & 80.7 & 2380 $\times$ 3 \\
        ViViT-L/16 \cite{arnab2021vivit} & 82.8 & 1446 $\times$ 12 \\
        \midrule
        Swin-L \cite{liu2021video} & 83.1 & 604 $\times$ 12 \\
        Swin-L (384) \cite{liu2021video} & 84.6 & 2107 $\times$ 12 \\
        Swin-L (384) \cite{liu2021video} & 84.9 & 2107 $\times$ 50 \\
        \midrule
        TokenLearner 16at12 (L/16) & 82.1 & 766 $\times$ 6 \\
        TokenLearner 8at18 (L/16) & 83.2 & 1105 $\times$ 6 \\
        TokenLearner 16at12 (L/16) & 83.5 & 766 $\times$ 12 \\
        TokenLearner 8at18 (L/16) & 84.5 & 1105 $\times$ 12 \\
        \midrule
        TokenLearner 16at18 (L/14) & 84.7 & 1621 $\times$ 12 \\
        TokenLearner 16at18 (L/10) & \textbf{85.4} & 4076 $\times$ 12 \\
        \bottomrule
  \end{tabular}
\end{table}

\subsection{Results}

We evaluate various versions of the ViT-L models incorporating the TokenLearner module. As mentioned above, all of the models are pre-trained on JFT and finetuned on Kinetics. 
In addition to the standard L/16 models + TokenLearner, we use the L/14 and L/10 models introduced in Tables \ref{table-large} and \ref{table-sota}. These models use additional layers compared to the standard ViT L/16, but as also described in the previous sections, the computation increase caused by them are minimal due to the number of tokens being much smaller, 8 or 16, in the added layers. We report both their classification accuracies and the computation in FLOPS.

Table~\ref{table-vivit-kin400} compares the accuracies of the base ViViT models against our ViViT + TokenLearner models on Kinetics-400. These models are directly comparable as they follow the exact same setting and the pre-train dataset. ``TokenLearner 16at12'' means that we have the TokenLearner layer learning 16 tokens, after the 12th Transformer layer. We are able to observe that the use of TokenLearner enables a better classification accuracy while also reducing the compute. Table~\ref{table-vivit-kin-sota} compares the TokenLearner accuracy against the prior models. Note that these approaches follow slightly different settings and pretrain datasets (e.g., the use of ImageNet-21K instead of JFT like ours). We believe the accuracy of 85.4 is the highest that has been reported so far, and we believe it is meaningful.

Table~\ref{table-vivit-kin600} compares the results on Kinetics-600. Similar to our results on Kinetics-400, when TokenLearner was first released, it extended the state-of-the-arts while also being computationally efficient.

\begin{table}
  \caption{ViViT + TokenLearner on Kinetics-600. The multiplication in GFLOPS correponds to the number of views used for the inference, such as 4x3 = 12. TL stands for TokenLearner.}
  \label{table-vivit-kin600}
  \centering
  \begin{tabular}{l|ccc}
        \toprule
        Method & Top-1 & Top-5 & total GFLOPS \\
        \midrule
        SlowFast 16x8, R101+NL \cite{feichtenhofer2018slowfast} & 81.8 & 95.1 & 234 $\times$ 30\\
        X3D-XL \cite{feichtenhofer2020x3d} & 81.9 & 95.5 & 48 $\times$ 30 \\
        TimeSformer-HR \cite{bertasius2021timesformer} & 82.4 & 96.0 & 1703 $\times$ 3 \\
        ViViT-L/16 \cite{arnab2021vivit} & 84.3 & 96.2 & 1446 $\times$ 12 \\
        \midrule
        Swin-B \cite{liu2021video} & 84.0 & 96.5 & 282 $\times$ 12 \\
        Swin-L (384) \cite{liu2021video} & 85.9 & 97.1 & 2107 $\times$ 12 \\
        Swin-L (384) \cite{liu2021video} & 86.1 & 97.3 & 2107 $\times$ 50 \\
        \midrule
        TL 16at12 (L/16) & 84.4 & 96.0 & 766 $\times$ 12 \\
        TL 8at18 (L/16) & 86.0 & 97.0 & 1105 $\times$ 12 \\
        \midrule
        TL 16at18 (L/10) & 86.1 & 97.0 & 4076 $\times$ 12 \\
        TL 16at18 w. Fuser (L/10) & \textbf{86.3} & 97.0 & 4100 $\times$ 12 \\
        \bottomrule
  \end{tabular}
\end{table}

\section{Experiments with Videos: TokenLearner with Bottleneck Transformer}

\subsection{Network architecture implementation}








In this experiment, we follow the Bottleneck Transformer~\cite{srinivas2021bottleneck} network style, while taking advantage of X3D~\cite{feichtenhofer2020x3d} as the backbone. This is motivated by the successful usage of X3D on Charades. Charades has longer duration videos (average of 30 seconds) with long actions (average action length of 15 seconds). This requires the model to understand longer term temporal information by considering multiple temporal tokens, and TokenLearner allows efficient computation of them over many frames.


Specifically, we modified X3D to be more computationally efficient by (1) replacing its 3D XYT convolutional layers with a pair of 2D conv. layer and 1D conv. layer, and (2) removing Squeeze-and-Excitation layers~\cite{hu2018squeeze} and swish activations. Our backbone could be viewed as X(2+1)D. We use the channel sizes and the number of layers identical to X3D-M, which is an efficient model.

\begin{figure}
    \centering
    \includegraphics[width=0.4\linewidth]{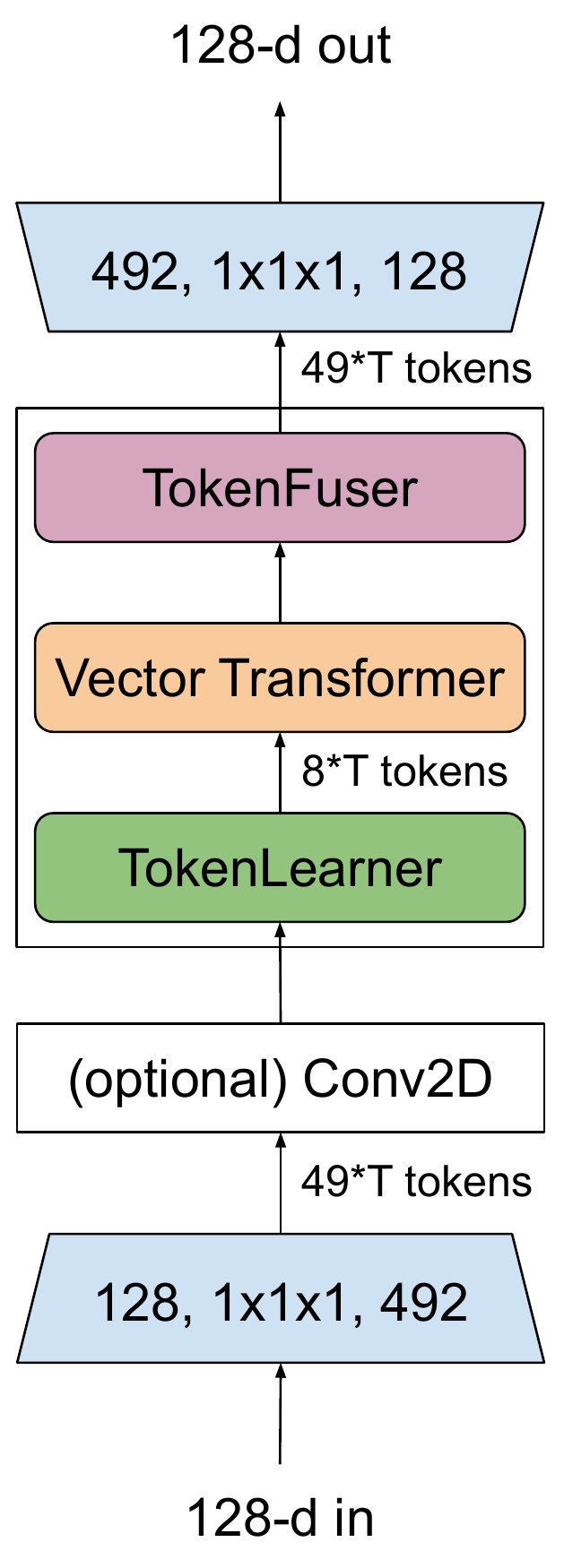}
    \caption{Our network module following the bottleneck transformer, with X(2+1)D backbone. It is an inverted bottleneck.}
    \label{fig:bottleneck}
\end{figure}

Based on such X(2+1)D architecture, and following the Bottleneck Transformer concept, we replace the space-time convolution layers in the last block with our transformers. 
Figure~\ref{fig:bottleneck} illustrates the residual module architecture, which is repeated multiple times in the block.
We have tried different versions, and our final model is built by replacing 1D temporal convolution with our TokenLearner while keeping the 2D $3\times3$ convolution layer in the X(2+1)D modules.
The spatial attention function (i.e., $\alpha(\cdot)$) in TokenLearner is implemented with a single conv2d layer.

Here, we used a Vector Transformer instead of MHSA as our Transformer layer, which could be also viewed as the MHSA with the number of heads being identical to the number of channels. We provide more details in Appendix.

We use $224\times224\times64$ videos for training and $256\times256\times64$ videos for testing. After the 3rd residual block, the input tensor has the shape of $8\times8\times64$, and this becomes the input to the TokenLearner. For an efficient implementation the intermediate channel size of TokenLearner was set identical to the output channel size, $d=432$. Notice that 64 frames were used to best capture longer-term temporal information. $S = 8$ number of tokens were used.




\subsubsection{Datasets}

\textbf{Charades dataset}: The Charades dataset~\cite{sigurdsson2016hollywood} is a dataset collected by assigning activity tasks which people in various environments are acting out, for example by performing a sequence of actions which involve interaction with objects. For example, sitting on the couch and reading a book, closing the book, standing up and speaking on the phone. It comprises $8000$ training and $1686$ validation videos with an average duration of $30$ seconds. It has 157 activity classes. This dataset is very challenging as it is a multi-class, multi-label video dataset, that is, more than one activity can occur at the same time, and it includes fine grained motions or interactions with small objects in real-world environments. We follow the standard evaluation protocols, reporting the mean Average Precision (mAP) \% (v1 classification setting of the dataset). We used the frame rate of 6 fps and 12 fps to obtain the training/testing videos (12 fps worked better). The dataset has a Non-Commercial Use license.

\textbf{AViD dataset}: The Anonymized Videos from Diverse countries (AViD) dataset~\cite{piergiovanni2020avid} is a unique dataset which is representative of the world's population video content generation. It is collected from videos uploaded from multiple countries across six continents and demonstrates higher diversity compared to other video datasets such as Kinetics in its concepts, actions and visual representations. For example a `greeting' in certain countries involves a handshake, in some a kiss, but in others a slight bow. The dataset is explicitly designed to contain less bias, encourage diversity, while respecting privacy and licenses.
The AViD dataset contains $887$ classes and $450$k videos ($410$k training $40$k testing) and is of comparable size to Kinetics-400 and Kinetics-600 datasets with $400$ and $600$ classes respectively, also containing variable duration videos $3-15s$.
We report classification accuracy over the $887$ classes.

All the videos in this dataset have the Creative Commons License.

\subsection{Results}
\textbf{Charades dataset results:}
In Table~\ref{table-charades} we compare the proposed TokenLearner to the state-of-the-art methods.
Our approach outperforms these, including recent work of which is most aligned to ours. The mAP of 66.3\% on Charades classification establishes the new state-of-the-art.


\begin{table}
  \caption{Performance on the Charades multi-label classification task. 12 fps setting. Performance is measured using the Mean Average Precision (mAP) since more than one ground truth action is possible. Methods with RGB and optical flow input modalities are listed.}
  \label{table-charades}
  \centering
  \begin{tabular}{lllc}
    \toprule
    Method & Input & Pre-train & ~~mAP~~\\
    \midrule
    I3D~\cite{carreira2017quo} & RGB & Kinetics &32.9\\
    I3D from~\cite{wang2018non} & RGB & Kinetics &35.5 \\
    I3D + Non-local~\cite{wang2018non} & RGB & Kinetics & 37.5 \\
    EvaNet~\cite{piergiovanni2018evolving} & RGB & Kinetics & 38.1\\
    STRG~\cite{wang2018videos} & RGB & Kinetics & 39.7 \\
    LFB-101~\cite{wu2018long} & RGB & Kinetics & 42.5 \\
    SGFB-101~\cite{ji2020genome} & RGB & Kinetics & 44.3 \\
    SlowFast-101~\cite{feichtenhofer2018slowfast} & RGB+RGB & Kinetics & 45.2 \\
    AssembleNet-50~\cite{ryoo2019assemblenet} & RGB+Flow & None & 47.0 \\
    Multiscale ViT~\cite{fan2021multiscale} & RGB & Kinetics & 47.7 \\
    AssembleNet-101~\cite{ryoo2019assemblenet} & RGB+Flow & Kinetics & 58.6 \\
    AssembleNet++~\cite{ryoo2020assemblenetplus} (w/o obj.) & RGB+Flow & None & 55.0\\  
    MoViNets~\cite{kondratyuk2021movinets} & RGB & None & 63.2\\
    \midrule
    Backbone (X(2+1)D-M) & RGB & None  & 62.7 \\
    Ours & RGB & None  & \textbf{66.3} \\

    \bottomrule
  \end{tabular}
\end{table}

\textbf{Anonymized Videos from Diverse countries (AViD) results:}
Table~\ref{table-avid} shows the results on the AViD dataset. As seen, our approach outperforms prior work on this challenging dataset too. We also compared ours to the reimplementation of TimeSformer module~\cite{bertasius2021timesformer} applied to the same backbone as ours.
This uses disjoint spatial and temporal transformer modules, which was also tested in \cite{arnab2021vivit}.
We are able to observe that we establish the new state-of-the-arts on this dataset, while also being more computationally efficient.
 
\begin{table}
  \caption{Performance on the Anonymized Videos from Diverse countries (AViD) dataset. Performance in terms of mean accuracy is shown in \% averaged over $887$ classes. Previous approaches results are reported from ~\cite{piergiovanni2020avid}, all based on training from scratch with RGB-only inputs. X(2+1)D-M baseline is with disjoint space+time Transformer (as in~\cite{bertasius2021timesformer}).}
  \label{table-avid}
  \centering
  \begin{tabular}{lll}
    \toprule
    Method & Accuracy & Total GFLOPS \\
    \midrule
    I3D~\cite{carreira2017quo} & 46.5 & 108 $\times$ N/A \\
    (2+1)D ResNet-50 & 46.7 & 152 $\times$ 115\\
    3D ResNet-50 & 47.9 & N/A\\
    SlowFast-50 8x8~\cite{feichtenhofer2018slowfast}  & 50.2  & 65.7 $\times$ 30 \\
    SlowFast-101 16x4~\cite{feichtenhofer2018slowfast}  & 50.8 & 213 $\times$ 30 \\
    \midrule
    Backbone (X(2+1)D-M)   & 48.6 & 532 $\times$ 1 \\
    X(2+1)D-M   & 50.6 & 493 $\times$ 1 \\
    Ours   &\textbf{53.8} & 487 $\times$ 1 \\
    \bottomrule
  \end{tabular}
\end{table}

\subsection{Ablations}

\subsubsection{Comparison against different tokenizations}


Here, we compare the model with TokenLearner against different space-time transformer tokenization strategies.
More specifically, we compare the use of TokenLearner + Vector Transformer + TokenFuser against the tokenization in the full joint space-time transformer module (advocated in \cite{arnab2021vivit} and also mentioned in \cite{bertasius2021timesformer}).
The full joint space-time transformer module is a transformer layer on space-time tokens similar to ours, but it relies only on the hand-designed tokenization. Compared to TokenLearner which generates $S \times T$ number of tokens, the full joint space-time transformer uses $H\times W \times T$ number of tokens. In our bottleneck implementation, it uses $\sim$8 times more tokens (i.e., 8*64 vs. 8*8*64). For the joint space-time transformer modules, the standard multi-head self-attention (MHSA) with 8 heads is used.


Table \ref{table-ablation-modules2} shows the results. Interestingly, despite the heavier computation of the full joint space-time transformer, it performed slightly worse to the TokenLearner modules. We believe this shows the advantage of the `adaptiveness' of the tokens in the TokenLearner and shows that the standard transformers might be suffering from the tokens irrelevant to the actions serving as noise or distractors.

We also report the amount of computation and the number of parameters of each module in these models. This depends on the input size and the hyper parameter setting, and our measurement is based on the input size (i.e., $T \times H \times W \times C$) of $8\times8\times64\times492$. Note that this is the measurement of modules, not the entire network.

\begin{table*}
    \caption{Comparison between TokenLearner and the joint space-time transformer modules similar to \cite{arnab2021vivit}, applied to our backbone. They use the X(2+1)D backbone, tested on Charades with the 6 fps setting, Charades 12 fps setting, and AViD dataset. GFLOPs and \# params are of each module (with 64 frame inputs), not the entire network.}
    \label{table-ablation-modules2}
    \centering
    \begin{tabular}{l|ccc|cc}
    \toprule
    Module & Charades-6fps & Charades-12fps & ~~AViD~~ & GFLOPs & \# params \\
    \midrule
    Joint space-time MHSA & 57.9 & 64.0 & 53.3 & 22.0 & 0.30M\\
    Conv2D + Joint space-time MHSA & 58.6 & 62.5 & 52.5 & 35.8 & 1.98M\\
    Ours (TokenLearner) & 58.8 & 63.4 & \textbf{53.8} & 3.4 & 0.81M\\
    Ours (Conv2D + TokenLearner) & \textbf{59.6} & \textbf{66.3} & 53.7 & 17.2 & 2.49M\\
    \bottomrule
    \end{tabular}
\end{table*}

\subsubsection{Comparison between multiple space-time layer combinations}

As also suggested in previous literature, it is a common strategy for video representations to pair a layer focusing on spatial information with a layer focusing on temporal information (e.g., R(2+1)D~\cite{tran2018closer} and TimeSformer~\cite{bertasius2021timesformer}). Table~\ref{table-ablation-modules} shows the results of this ablation. For spatial and temporal transformer implementations, the standard multi-head self-attention was used, as was done in \cite{arnab2021vivit,bertasius2021timesformer}. The result shows that the proposed TokenLearner is more accurate than other popular combinations. The modules based on TokenLearner also effectively only uses a fraction of the Tokens per frame (i.e., 8) as opposed to other methods which use $16\times16$ or $32\times32$ tokens.


One of the main benefits of the TokenLearner (in addition to the adaptive tokenization of the input and that we explicitly fuse the tokens to capture their spatio-temporal patterns) is that, unlike the disjoint space/time transformers used in this ablation study, it is a joint space-time transformer. Simultaneously, it still manages its computation to be much more tractable (as shown in Table~\ref{table-ablation-modules}): A naive full version of the space-time transformer would require consideration of $8\times8\times64 = 4096$ tokens in our case, building and multiply the attention tensor of size $4096\times4096$. On the other hand, the TokenLearner learns to consider $8\times64=512$ tokens jointly.



\begin{table}
    \caption{Comparison between different space-time transformer modules. They were all applied to the same backbone architecture (i.e., the Bottleneck Transformer-style with X(2+1)D). The Charades-6fps is used in this experiment. FLOPS are estimated with 64-frame settings, per module.}
    \label{table-ablation-modules}
    \centering
    \begin{tabular}{lccc}
    \toprule
    Module & Acc. (\%) & GFLOPs & \# params \\
    \midrule
    Conv2D + Conv1D & 56.6 & 18.3 & 2.24M\\
    Conv2D + MLPMixer~\cite{tolstikhin2021mlpmixer} & 57.0 & 13.8 & 2.06M\\
    Conv2D + Temporal transformer & 58.4 & 16.5 & 1.98M\\
    Spatial + Temporal transformer & 58.8 & 5.5 & 0.59M\\
    Conv2D + Spatial + Temporal  & 58.0 & 19.2 & 2.27M\\
    Ours (TokenLearner) & 58.8 & 3.4 & 0.81M\\
    Ours (SpatialT + TokenLearner) & 58.9 & 6.2 & 1.11M\\
    Ours (Conv2D + TokenLearner) & 59.6 & 17.2 & 2.49M\\
    \bottomrule
    \end{tabular}
\end{table}






\subsubsection{More TokenLearner alternatives}

We also compared our spatial attention-based token learning with alternative approaches: (1) using a fixed grid to split each frame into the same number of tokens (i.e., 8 tokens), (2) the approach of directly generating tokens using a fully connected layer, and (3) the approach of spatially average pooling the entire frame pixels and using fully connected layers to generate multiple tokens per frame. In the second approach, we directly model $z_i = A_i(x)$ as a dense layer, producing $T \times S \times C$ tensor based on the $T \times H \times W \times C$ input. The third approach is similar, except that we apply spatial global average pooling per frame and then use MLP to generate tokens.

The fixed split tokenization method (1) provided us the accuracy of 58.8 on Charades, as opposed to 59.6 of ours. The direct token generation method (2) provided the accuracy of 56.6 on Charades, failing to obtain better tokens. Pooling and generation method (3) gave us the accuracy of 58.6. These results suggest the importance of spatial attention for the token learning, our TokenLearner. The same vector transformer and TokenFuser (from Section~\ref{sec:main}) were used for this ablation.



\begin{figure*}
    \centering
    \includegraphics[width=0.8\textwidth]{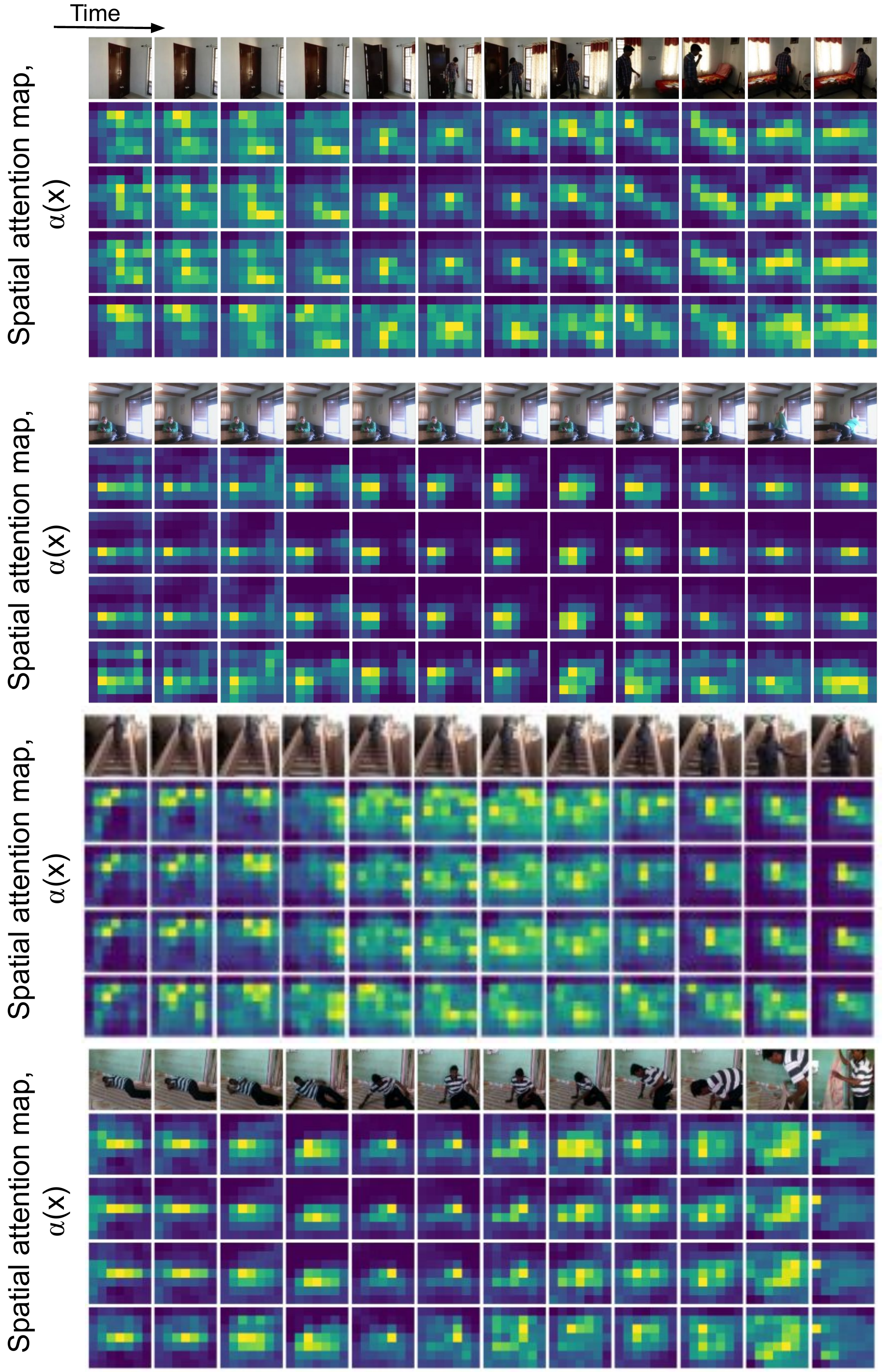}
    \caption{Visualization of the spatial attention maps for the tokenizations. Attention maps for four among a total of eight learned tokens are shown.}
    \label{fig:visualization}
\end{figure*}

\subsection{Visualizations}

Figure~\ref{fig:visualization} shows visualizations of the tokens being learned with our approach. We show the spatial attention maps (i.e., $\alpha_{i}(x)$) from the first TokenLearner module, as the inputs to the higher-level TokenLearner becomes more mixed spatially and temporally. We are able to observe that they tend to focus more on human regions, and that they change over time responding to the changes in the visual input. Among the $S=8$ tokens per frame we learn, we visualize 4 of them.


\section{Related work}

Visual understanding is a long-standing problem in computer vision. Image recognition tasks including object classification and detection have been extensively studied in many different directions. Most of today's methods focus on learning to represent spatial appearance in images. It is a challenging task which spans many years of computer vision research~\cite{rawat2017deep}.

Video understanding has an even more challenging task for extracting both the spatial and the temporal information in a video~\cite{carreira2017quo,zhou2018temporal,tran2018closer,wu2021towards,feichtenhofer2016convolutional,timeception,bertasius2021timesformer,korbar2019scsampler,eco2018}. 
In order to adequately capture both motion and appearance information in videos, full 3D space-time convolutional layers as well as (2+1)D convolutional layers have been used~\cite{tran2014c3d,carreira2017quo,tran2018closer,xie2018rethinking}. 
More advanced network designs have also been extremely popular in video CNNs particularly two-stream ones~\cite{simonyan2014two,feichtenhofer2016tsres,feichtenhofer2017tsmult,feichtenhofer2016convolutional,diba2019holistic,feichtenhofer2018slowfast} and, recently, architecture searched ones~\cite{feichtenhofer2020x3d,ryoo2019assemblenet,ryoo2020assemblenetplus,piergiovanni2018evolving,piergiovanni2021tvn,ViPNAS}. 

Attention-based architectures, e.g., the Transformer~\cite{vaswani2017attention} have shown remarkable success in both Natural Language processing and Computer Vision. 
The Vision Transformer~\cite{dosovitskiy2020} demonstrated how the NLP-specific Transformer architecture can elegantly work for images, and image recognition tasks. This is done by subdividing the input image into non-overlapping patches on  a  regular grid and feeding them as token embeddings to the Trasnformer, where $O(N^2)$ tokens are used or order of 256 or 1024. A plethora of approaches have followed this strategy~\cite{touvon2021going,levit,yuan2021t2t,zhai2021scaling,fan2021multiscale}, with some of the approaches proposing multi-scale visual transformer versions~\cite{fan2021multiscale,levit,pyramidVIT}. Some methods focus on optimizations of these models and layers, and they also have been successful~\cite{carion2020detr,zhao2020vectorattention,cordonnier2020on,ramachandran2019sasa}.

Applying attention-based architectures to videos presents a definite challenge as the model needs to learn dependencies across both the spatial and temporal domains.
\cite{girdhar2019video} relied on the region proposal network to use the detected human and object candidates as tokens, showing that it could be combined with CNNs.
A couple of recent works \cite{arnab2021vivit,bertasius2021timesformer}, in the spirit of the Vision Transformer, subdivided the video into token in a 3D grid to capture the video input. This leads to $O(N^3)$ increase in the number of tokens required for learning (typically $\sim25$k tokens for 96-frame model).
Attention-based architectures have also been used in the context of video generation~\cite{weissenborn2020videotransformer}.
Several architectures have demonstrated attention-based architectures for handling multiple inputs of various modalities~\cite{perceiver,lee2019set}. 

Our work, in contrast to both related work in image and video recognition, learns the tokens from data which results in a significantly fewer tokens, and more efficient approach. We see that even 8x times fewer tokens (e.g., 512 vs 4096), when learned, are able to capture successfully the information needed for video representation learning.
Importantly, our proposed TokenLearner, is applicable to both video and image recognition tasks achieving better results in both domains.

\section{Conclusions}

We have presented TokenLearner, a novel approach for visual representation learning, which adaptively tokenizes the inputs. The goal is to learn to extract important tokens in images and video frames for the recognition tasks at hand. Our approach is more efficient, than contemporary work, by finding few important space-time tokens which can model visual representations of images and videos. We observe improved accuracies across image classification and challenging video understanding tasks, and outperformed prior approaches in many datasets. One of the remaining challenges is in learning full spatio-temporal tokens. The current TokenLearner focuses on finding spatial tokens over a sequence of frames, and it could be extended to mine tokens over space-time volumes.

\section*{Acknowledgement}

We thank Dmitry Kalashnikov, Andy Zeng, and Robotics at Google NYC team members for valuable discussions on attention mechanisms.




\appendices

\section{Vector Transformer: Pairwise vector attention}
\label{sec:pw_att}

Here, we summarize the details of the Vector Transformer used in the Bottleneck Transformer experiments.

Once TokenLearner generates adaptively learned tokens, a vector attention between key-query pairs could be computed. This can be thought as a version of multi-head self-attention in which the number of heads is the same as channels, allowing us to learn a different attention matrix for each channel. It captures in an efficient way pairwise space-time relations per channel, particularly benefiting tokens with rich channel information.

Given $Z$, a set of tokens reflecting different space-time aspects of a video, the Transformer models space-time interactions between them. In particular, we follow the formulation of~\cite{zhao2020vectorattention}, which enables a vector-version of the Transformer, although it is also possible to incorporate other attention mechanisms.

For every token $z_i$, the output of the Transformer $y_i$ is computed by considering all possible $z_j$ as:
\begin{equation}
y_i = \sum_{z_j \in Z} \gamma(f_q(z_i) \odot f_k(z_j)) \odot f_v(z_j)
\end{equation}
where $i$ and $j$ are the indexes of the tokens in $Z$ whose size is $|Z| = ST$. $f_q$, $f_k$, and $f_v$ are the linear layers projecting the tokens. $\gamma$ is an extra projection layer to match the channel dimensionality followed by a softmax function over $j$. When the channel sizes of the projections are identical, $\gamma$ is simplified as a single softmax layer identical to the standard transformer.


In the original transformer notation, the query matrix $Q$ corresponds to our $\{f_q(z_i)\}_i$, and the key matrix $K$ corresponds to our $\{f_k(z_j)\}_j$. Instead of computing the dot product between $Q$ and $K$ as $QK^T$ to generate the attention `matrix', this vector formulation computes an attention `tensor' $\{\gamma(f_q(z_i) \odot f_k(z_j))\}_{(i, j)}$ preserving the channel information. It has shape $ST \times ST \times d$ where $d$ is the intermediate channel size. The computed attention tensor is multiplied with the value matrix $\{f_v(z_j)\}_j$ to get the final transformer outputs.

Notice that this vector transformer is a global representation, and the temporal range of the information it is able to capture entirely depends on what tokens we provide to it. With our learnable adaptive tokens, we have the capability to cover a larger number of frames and focus on the temporal structure.


\section{Image and Video Classification Training details}

\subsection{Image classification}
We follow the exact training protocols and the hyper parameters of \cite{dosovitskiy2020} for our image classification experiments.

\subsection{Video classification - ViViT}

We follow the exact training protocols and the hyper parameters of \cite{arnab2021vivit}. We use the same code (the Scenic library~\cite{dehghani2021scenic}) and the hardware for the training as well as for the evaluation.

We train the Kinetics model for 30 epochs with the base learning rate of 0.05 with the Momentum optimizer. Basically, all the settings in our Kinetics experiments follow the setting of ViViT.

\subsection{Video classification - Bottleneck Transformer}

We provide the training details as below. For the training/testing splits of the datasets, we followed their standard settings.

We use the cosine-decay learning rate which was popularly used in many video CNN model trainings. The base learning rate of 0.8 per TPU core (which is equivalent to a single GPU) is used for the Charades dataset (multi-label action classification) and the base rate of 0.025 per TPU was used for the AViD dataset (video classification). The training was done for 100k iterations with the batch size of 4 per TPU core (i.e., 4*64=256 was our batch size) in the Charades experiments. The batch size of 8 per TPU core was used for AViD. 100k iterations correspond to roughly 125 epoches in AViD. Label smoothing of 0.2 was used for the AViD training. No label smoothing was used for the Charades. In Charades, the training was done by temporally cropping a long Charades videos (e.g., 30 seconds) into 64 frame segments. The evaluation was done similarly with 64 frame segments by merging their output responses.

The training time of a single model was around $\sim$16 hours with 32 TPU v3.
This was bottlenecked by the data pipeline, and the actual computation is less.

\section{Comparing more models on few-shot learning}

Here, we report few-shot learning accuracy of more models with TokenLearner, extending Figure \ref{fig:fewshot}. Specifically, we additionally show L/16 (and L/14) models with TokenLearner at various locations, including inserting it at the 2nd, 3rd, 6th, and 12th attention layers. We are able to observe the strategy of adding TokenLearner early in the network allows us to save the computation while obtain superior image classification accuracy to the base models.

\begin{figure}
    \centering
    \includegraphics[width=0.49\linewidth]{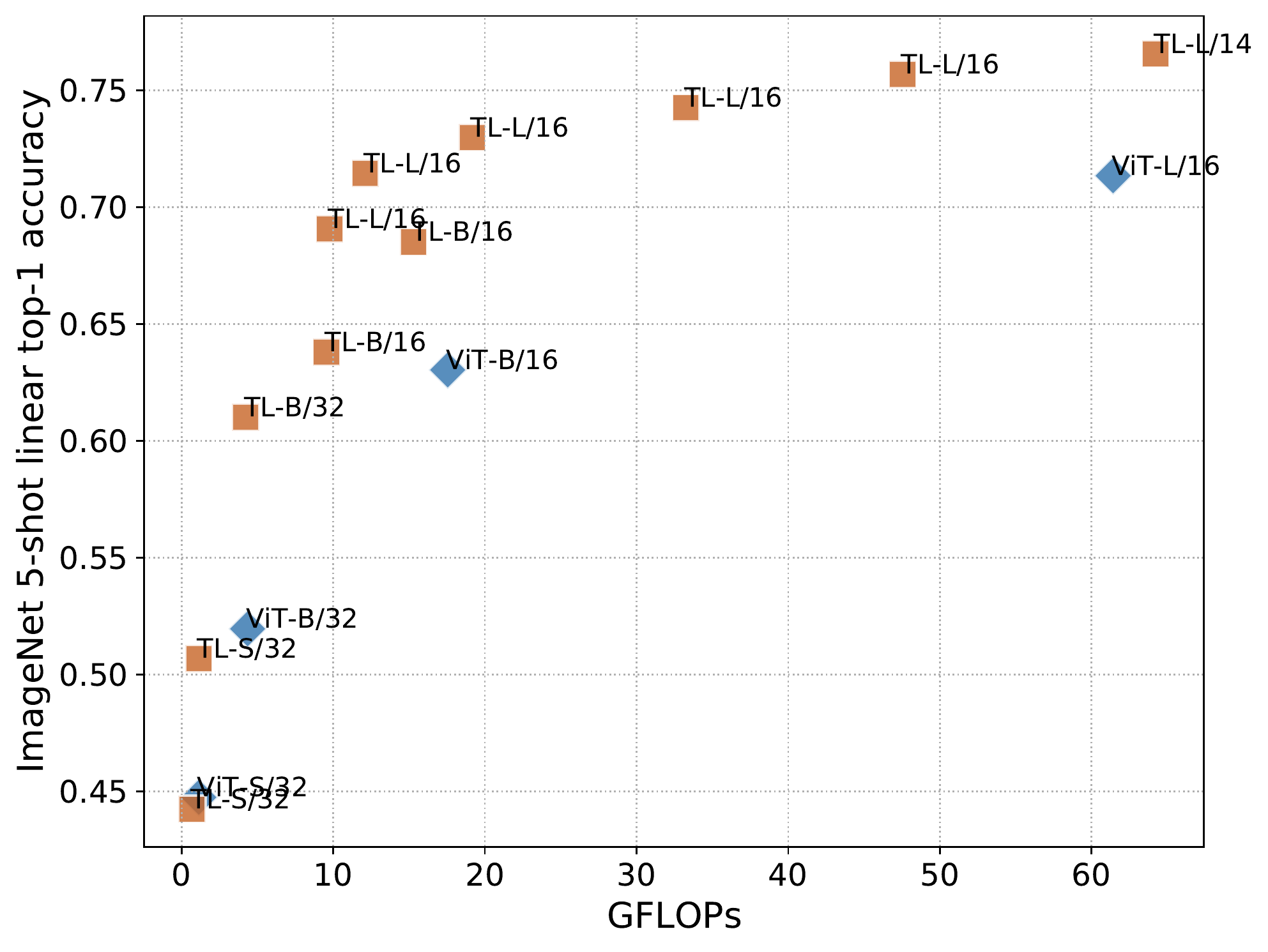}
    \includegraphics[width=0.49\linewidth]{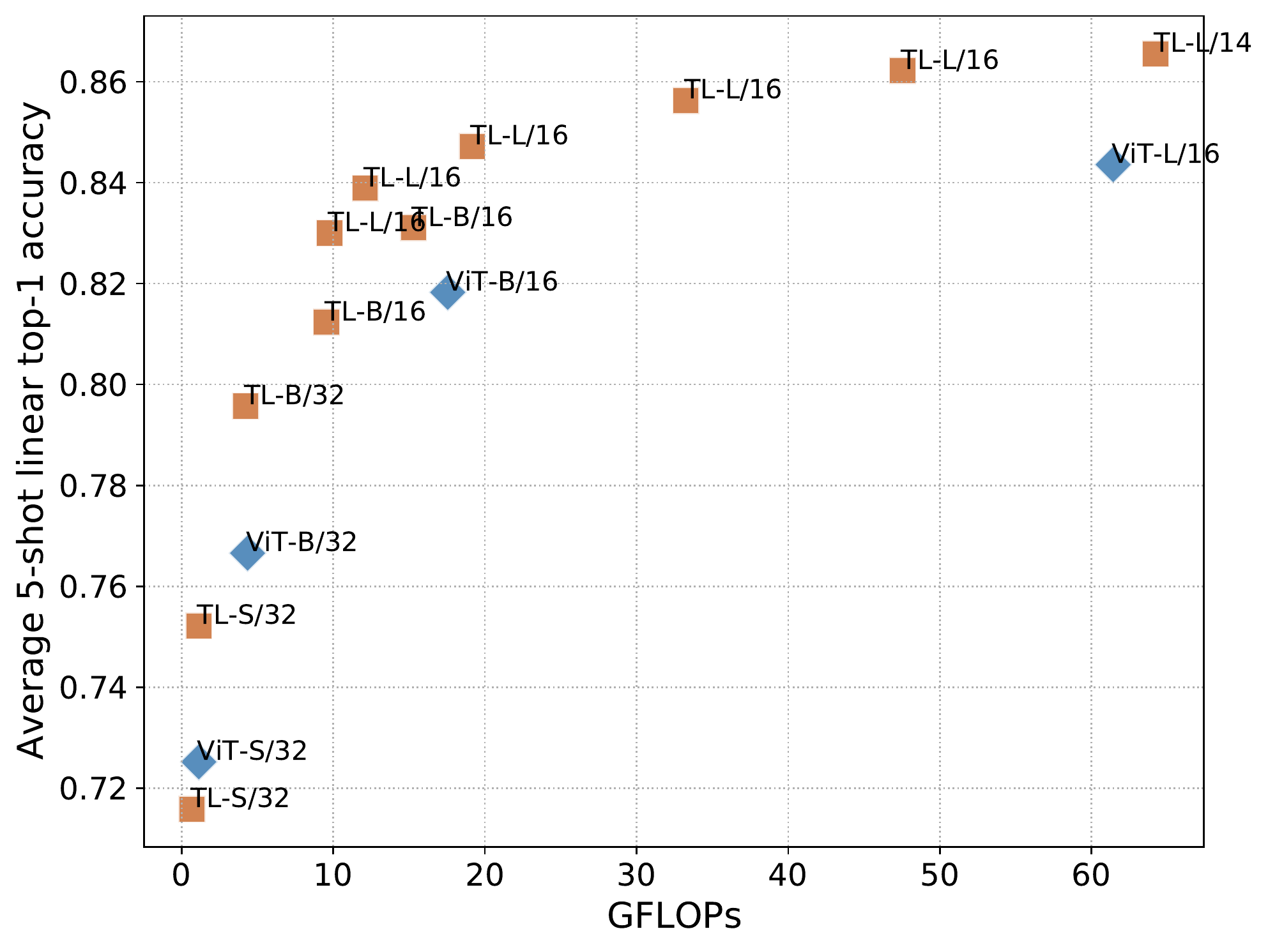}
    \caption{Few-shot classification experiments. It shows 5-shot classification accuracies on ImageNet (left) and average of multiple datasets listed in Sec.~\ref{sec:exp_image_results} (right).
    `TL' stands for TokenLearner.}
    \label{fig:fewshot-more}
\end{figure}

\section{Additional ablations}

\subsection{Different components}

Using the setting of the Bottleneck Transformer experiments, we did an ablation to evaluate components of our approach and their combinations. We conducted ablations removing/adding different components of our model. In addition to Vector Transformer described in the above subsection, we also tried an ablation of replacing it with the multi-head self-attention.  Table~\ref{table-ablation-components} shows the results, demonstrating the benefits each of the elements bring to the approach. For this experiment, we used the module composed of Conv2D + transformer (within the bottleneck), which we found to perform the best from the other ablations.

\begin{table}
    \caption{Comparing different components of our TokenLearner. On Charades dataset (6fps).}
    \label{table-ablation-components}
    \centering
    \begin{tabular}{lcc}
    \toprule
    Module & Accuracy (\%) \\
    \midrule
    Standard transformer (MHSA) & 58.4 \\
    Vector transformer (VectT) & 58.1 \\
    Prior-only-attention + broadcasting & 58.6\\
    Vector transformer (VectT) + broadcasting & 58.9\\
    Vector transformer (VectT) + TokenFuser & 59.0\\
    TokenLearner + MHSA + TokenFuser & 59.0\\
    TokenLearner + VectT + TokenFuser & 59.6\\
    \bottomrule
    \end{tabular}
\end{table}

{
\small

\bibliographystyle{IEEEtran}
\bibliography{bib}
}

\begin{IEEEbiographynophoto}{Michael S. Ryoo} is a SUNY Empire Innovation Associate Professor in the Department of Computer Science at Stony Brook University, and is also a staff research scientist at Robotics at Google. He previously was an assistant professor at Indiana University Bloomington, and was a staff researcher within the Robotics Section of NASA's Jet Propulsion Laboratory (JPL). Dr. Ryoo received his Ph.D. from the University of Texas at Austin and B.S. from Korea Advanced Institute of Science and Technology (KAIST).
\end{IEEEbiographynophoto}

\begin{IEEEbiographynophoto}{AJ Piergiovanni} is a research scientist at Google. He has a PhD in computer science from Indiana University and a BS from Rose-Hulman Institute of Technology. His research interests are in video understanding, building efficient models and learning from vision and language.
\end{IEEEbiographynophoto}

\begin{IEEEbiographynophoto}{Anurag Arnab} is a research scientist at Google. Previously, he completed his PhD at the University of Oxford.
\end{IEEEbiographynophoto}

\begin{IEEEbiographynophoto}{Mostafa Dehghani} is a research scientist at Google Brain. Previously, he completed his PhD at the University of Amsterdam.
\end{IEEEbiographynophoto}

\begin{IEEEbiographynophoto}{Anelia Angelova} is a research scientist in the area of computer vision. She leads the Vision and Language team in Brain Research at Google and was previously leading the Robot Vision team in Robotics at Google. Her research interests span many topics in computer vision: object recognition and detection, 3D scene understanding, self-supervised learning, video understanding, multi-modal learning, robotics perception, real-time algorithms and others. She has integrated her work in production systems, including X Robotics, Google Maps, Google Cloud, and Google's self-driving car (Waymo). Anelia received her MS and PhD degrees in Computer Science from California Institute of Technology.
\end{IEEEbiographynophoto}

\end{document}